\documentclass[10pt]{article}
\usepackage[a4paper,margin=20mm]{geometry}
\usepackage[T1]{fontenc}
\usepackage[utf8]{inputenc}
\usepackage{lmodern}
\usepackage{xcolor}
\usepackage{colortbl}
\usepackage{titlesec}
\usepackage[most]{tcolorbox}
\usepackage{amsmath,amssymb}
\usepackage{longtable,booktabs,array,multirow,calc}
\usepackage{capt-of}
\usepackage{caption}
\usepackage{enumitem}
\usepackage{graphicx}
\usepackage{fancyhdr}
\usepackage[hidelinks]{hyperref}
\usepackage{placeins}

\let\oldsection\section
\renewcommand{\section}{\FloatBarrier\oldsection}
\let\oldsubsection\subsection
\renewcommand{\subsection}{\FloatBarrier\oldsubsection}
\definecolor{awObsidian}{HTML}{0D0B1F}
\definecolor{awMist}{HTML}{C7D2E4}
\definecolor{awIndigo}{HTML}{3B4A66}
\definecolor{awViolet}{HTML}{A24BF5}

\hypersetup{
  linkcolor=awViolet, citecolor=awIndigo, urlcolor=awViolet,
  pdftitle={AutoE2E: Don't Speak -- Language-Free End-to-End Autonomy},
  pdfauthor={Autoware Foundation -- Robotaxi Working Group}
}

\titleformat{\section}
  {\color{awObsidian}\normalfont\Large\bfseries}
  {\thesection}{0.6em}{}
  [{\color{awViolet}\titlerule[1pt]}]
\titleformat{\subsection}
  {\color{awIndigo}\normalfont\large\bfseries}
  {\thesubsection}{0.6em}{}

\renewcommand{\headrulewidth}{0.4pt}
\renewcommand{\headrule}{\hbox to\headwidth{\color{awMist}\leaders\hrule height \headrulewidth\hfill}}
\newtcolorbox{partnerbox}[1]{
  enhanced, breakable, colback=white, colframe=awIndigo,
  coltitle=white, colbacktitle=awIndigo, fonttitle=\bfseries,
  title={#1}, boxrule=0.8pt, arc=1.5mm, left=7pt, right=7pt, top=3pt, bottom=3pt
}

\begin{document}

\noindent
\begin{minipage}[t]{0.30\textwidth}
\vspace{0pt}
\includegraphics[width=\linewidth]{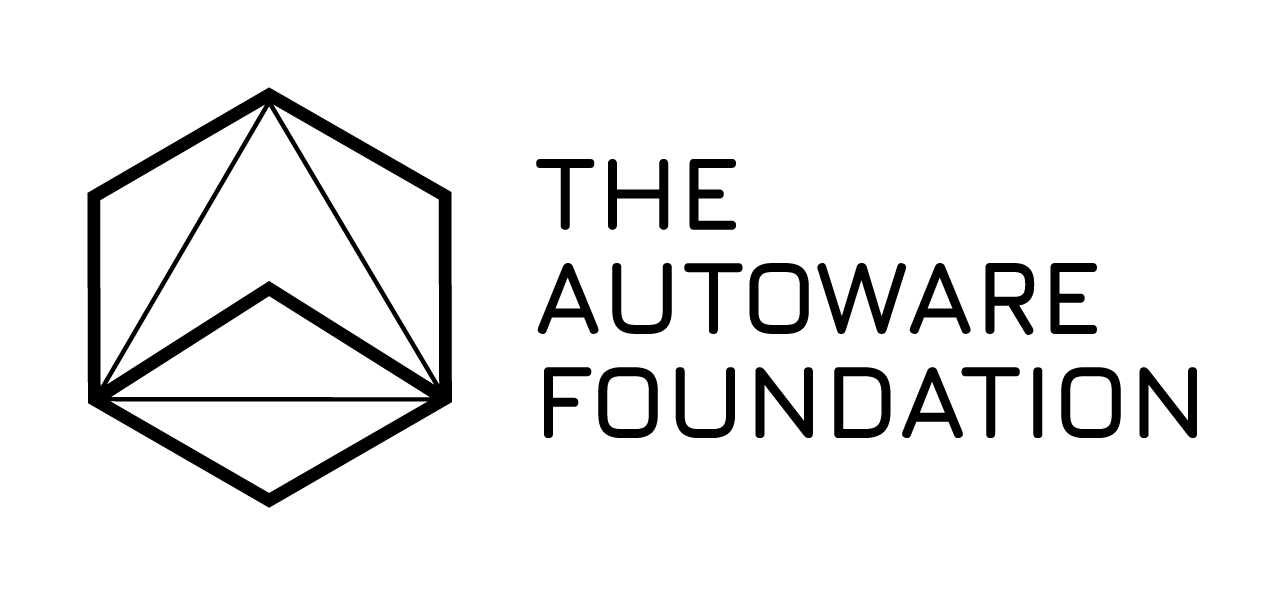}
\end{minipage}%

\vspace{0.6em}
{\Large\bfseries\color{awObsidian}AutoE2E: Don't speak - Language free end-to-end autonomy}\\[2pt]
\hspace{2.2cm}{\large\color{awViolet}\quad \quad for L2++ ADAS and L4 Robotaxis}

\vspace{4pt}
\noindent{\color{awViolet}\rule{\linewidth}{1.4pt}}

\vspace{0.8em}
\noindent Autoware Foundation Robotaxi Working Group\footnote{A full
contributor list is given in Sec.~\ref{sec:contributors}.}

\vspace{1em}

\begin{abstract}
AutoE2E is an open-source end-to-end driving model developed by the Autoware Foundation Robotaxi Working Group. It introduces a temporal bird's-eye-view (BEV) architecture that combines multi-camera perception, HD-map information and route conditioning to predict future vehicle motion. 
The evaluated model uses a pretrained camera-BEV backbone, temporal feature fusion and a lightweight recurrent trajectory planner. It contains approximately 79.9 million parameters, of which approximately 1.45 million are trainable in the evaluated configuration.
The v1.0 release provides model checkpoints, training and evaluation code, and DataModelConsole, a dashboard for visualizing and inspecting model experiments. On the KITScenes validation benchmark, the fine-tuned checkpoint achieves an average displacement error (ADE) of 0.745 m and a final displacement error (FDE) of 2.020 m at a three-second prediction horizon.
The paper also reports preliminary real-vehicle research conducted with the Technical University of Munich's EDGAR platform, alongside ongoing collaboration with the Karlsruhe Institute of Technology.
AutoE2E v1.0 is an early open research release. Its quantitative evaluation is open-loop, and the model has not yet demonstrated validated closed-loop autonomous driving. The release is intended to provide the autonomous-driving community with an inspectable baseline for further research, experimentation and collaborative development.
\end{abstract}

\noindent\textbf{Code and release:} \url{https://github.com/autowarefoundation/auto_e2e/releases/tag/v1.0} (Apache-2.0) \\
\textbf{Checkpoints:} \url{https://huggingface.co/AutowareFoundation/auto_e2e} (\texttt{kitscenes-epoch-5}, fine-tuned; \texttt{nuplan-epoch-5}, baseline -- weight license follows the source training datasets.\\
\textbf{Dashboard:} \url{https://d2itskdqq39tx1.cloudfront.net/} (DataModelConsole, Sec.~\ref{sec:dmc})


\tableofcontents

\section{Introduction}\label{introduction}

\begin{figure}[h]
\centering
\includegraphics[width=\linewidth]{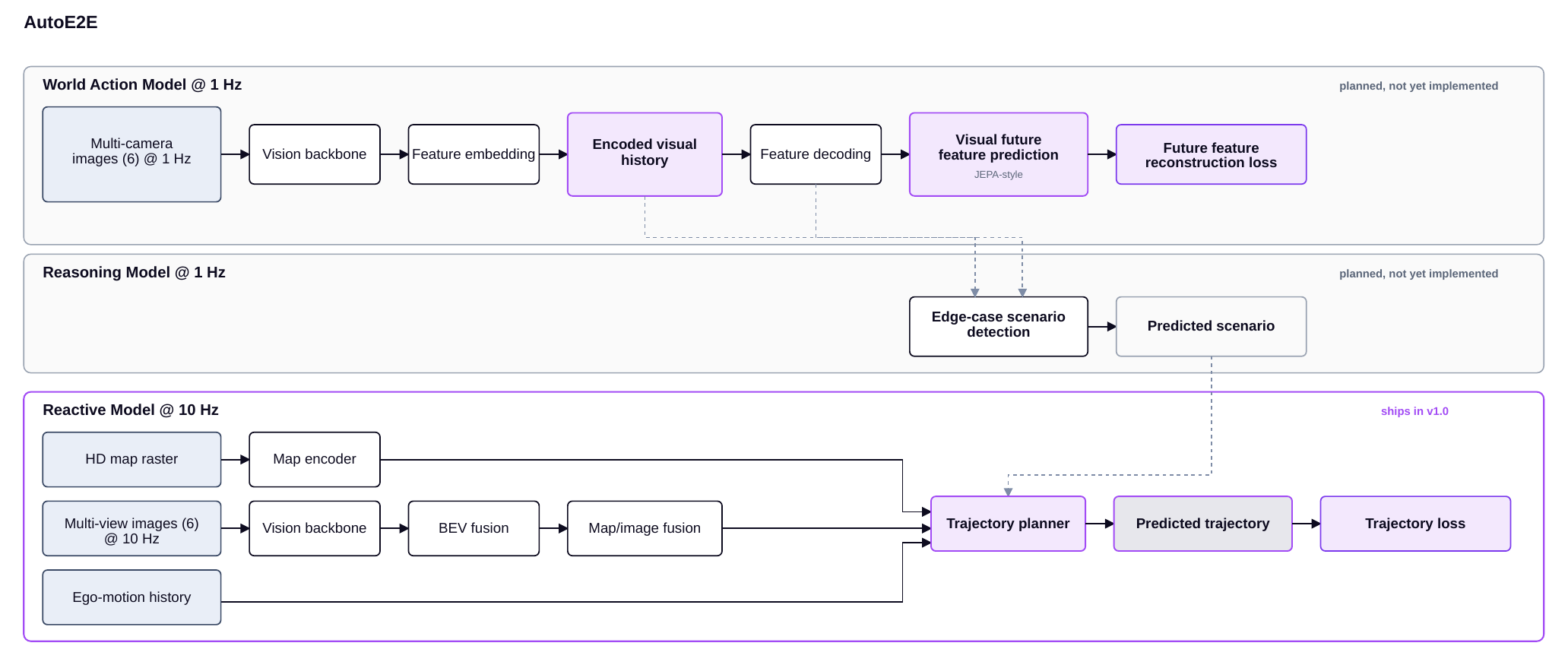}
\caption{The working group's planned three-band architecture. The
reactive model at 10\,Hz (violet band) is AutoE2E, the subject of this
paper, already implemented and evaluated here. The world-action model
at 1\,Hz and the reasoning model at 1\,Hz (grey bands) are planned
interfaces, not yet implemented or trained; no result in this paper
depends on them. The world-action band's objective is explicitly
non-linguistic: a JEPA-style future-feature reconstruction loss in the
camera's own embedding space, not a language-model reasoning trace --
the design choice the paper's title refers to.}\label{fig-threeband}
\end{figure}

The Autoware Foundation's Robotaxi Working Group is developing an
open-source, three-band architecture for autonomous driving
(Fig.~\ref{fig-threeband}): a reactive band that maps sensing directly
to control, a reasoning band, and a world-action band. The three bands
are intended to be interfaced together once each is mature, not trained
as one monolithic network; development starts with the reactive band,
the one expected to change fastest, and is designed so that reasoning
and world-action components can be interfaced onto it as they come
online. AutoE2E is that reactive band. It is designed to be compact and
inspectable rather than to lead on raw capability: a deterministic
policy -- it outputs one trajectory per input, with no sampling at
inference, unlike the generative vision-language-action (VLA) planners
it is benchmarked against -- with 79.9\,M total parameters (1.45\,M
trainable) that reaches up to 76\,FPS on a single consumer GPU
(Sec.~\ref{sec:realvehicle}), orders of magnitude smaller and faster
than those planners.

A reasoning trace is inspectable in a way a raw trajectory is not, but
only if the trace is shown to correspond to the action actually taken --
a check the VLA literature mostly assumes rather than runs. VLA
planners couple a vision encoder and a language-model backbone to
produce trajectories together with a reasoning trace, typically through
a three-stage recipe of frozen-encoder projector pretraining, supervised
fine-tuning on driving data with teacher-generated chain-of-thought
annotations, and a reinforcement post-training stage that aligns the
policy with closed-loop or preference reward. Representative systems in
this family include Qwen-Drive-1.0~\cite{ref-qwendrive}, which unifies
3D perception, driving VQA, and motion planning on a pretrained
vision-language model while leaving its weights untouched, and
Alpamayo~\cite{ref-alpamayo}, built on a reasoning-focused backbone and
intended primarily as a distillation teacher; the common bottleneck
across the family is the gap between research-scale capability and
on-vehicle inference budget, addressed almost universally by
teacher-student distillation rather than by training a deployable model
directly. AutoE2E sits at the opposite end of this design space: a
deterministic, non-generative model with a fixed camera-BEV backbone,
no language component, and no sampling at inference -- and where the
working group's roadmap does add a future-prediction capability
(Sec.~\ref{sec:future}), it is specified as a JEPA-style latent-feature
objective, predicting the camera pathway's own future embeddings,
rather than a token sequence a language model decodes. What AutoE2E
gains is a system whose every claim in this paper can be checked
against source code and a fixed input/output contract rather than
against a reasoning trace whose faithfulness is unverified; what it
gives up is the open-vocabulary instruction following and explicit
reasoning output the VLA family offers.

One methodological commitment runs through this paper and is worth
stating up front. End-to-end (E2E) driving research has developed
largely around camera-only open-loop evaluation on
nuScenes~\cite{ref-nuscenes}; UniAD~\cite{ref-uniad} and
VAD~\cite{ref-vad} demonstrated planning-oriented integrated stacks in
this setting, with temporal BEV representations such as
BEVFormer~\cite{ref-bevformer} providing a shared basis. Deployed
vehicle policies, however, often receive more than cameras. AutoE2E's
deployment contract supplies an HD-map raster and a selected-route
raster at every step in addition to camera images. Evaluating this
policy only on a camera-only benchmark reports behavior under a
condition that omits information the policy was designed to consume;
reporting only the map-and-route condition would conversely hide
behavior when navigation inputs are unavailable. These evaluations
answer different questions and are kept separate throughout this paper.

\subsection{Related Work}\label{related-work}

End-to-end driving policies share a common design space -- how much of
the perception stack to fold into one network, whether language is part
of the model, how temporal information is represented, how a map or
route reaches the policy, and how the result is evaluated -- and
AutoE2E's choices in that space only mean something read against it.

\textbf{Planning-oriented end-to-end driving.} UniAD~\cite{ref-uniad}
links detection, tracking, mapping, motion forecasting, occupancy
prediction, and planning through shared query interaction, so that
every task contributes to the final plan. VAD~\cite{ref-vad} keeps the
same chain but replaces the raster scene with a vectorized one to cut
its computational cost, and
TransFuser~\cite{ref-transfuser}\cite{ref-transfuser-cvpr} fuses camera
and LiDAR through a transformer and evaluates the result in closed-loop
CARLA, a different evaluation setting from the open-loop one this paper
uses. AutoE2E makes a narrower commitment: it does not reproduce the
detection/tracking/mapping/occupancy chain, only the idea of a shared
BEV representation feeding a planner, plus explicit navigation context
and auxiliary route reconstruction. It has not been compared against
UniAD on a common benchmark, and no claim of superiority or inferiority
is made.

\textbf{Temporal BEV representations.} How a temporal fusion strategy
folds history into the current frame determines what the resulting
representation can be attributed to: a given moment, a window, or an
unbounded past. BEVFormer~\cite{ref-bevformer} aggregates multi-camera
features into a BEV representation spatiotemporally, using learnable BEV
queries and deformable attention derived from Deformable
DETR~\cite{ref-deformable}. BEVFormer v2~\cite{ref-bevformerv2} adapts
modern image backbones to BEV perception through perspective
supervision, and releases an official ResNet-50 checkpoint together with
a temporal configuration that fuses multi-frame BEV by convolution -- an
explicit, bounded window rather than a recurrent state carried
indefinitely. AutoE2E's frozen camera pathway starts from this official
checkpoint (Sec.~\ref{sec:architecture}).

\textbf{Map- and route-conditioned planning.} ChauffeurNet~\cite{ref-chauffeurnet}
rendered the road map and the intended route as rasters for imitation
learning; VectorNet~\cite{ref-vectornet} and LaneGCN~\cite{ref-lanegcn}
encoded HD maps as vectors and lane graphs, showing that map structure
governs motion-forecasting accuracy. The UniAD-family baselines
evaluated on the KITScenes~\cite{ref-kitscenes} E2E benchmark are
conditioned on discrete navigation commands (turn left, turn right, go
straight); the route used here is instead a lane-sequence corridor over
a Lanelet2~\cite{ref-lanelet2} map with a destination marker, fitted
\emph{after the fact} to the trajectory the model is then scored
against (Sec.~\ref{sec:architecture}). That provenance is precisely why
route conditioning cannot be credited with an accuracy gain from this
protocol alone; isolating its effect requires a same-scene experiment
not yet run (Sec.~\ref{sec:limitations}).

\textbf{Evaluation protocols and benchmarks.} Open-loop planning
evaluation on nuScenes has repeatedly been shown to admit shortcuts by
which high scores are obtained without genuine driving
competence~\cite{ref-admlp,ref-egostatus}. Closed-loop evaluation in
nuPlan~\cite{ref-nuplan}, non-reactive simulation in
NAVSIM~\cite{ref-navsim}, the finding that rule-based planners can rival
learned ones under the standard protocol~\cite{ref-pdm}, and
conditioning-signal shortcut biases~\cite{ref-hidden} are different
instances of the same finding: change what the evaluation condition
reveals or withholds, and the ranking of models changes with it,
independent of what they actually learned. This paper remains in the
open-loop setting and adopts three design principles that follow from
this literature: align the evaluation setup with the deployment
contract, distinguish deterministic overlay replay (which does not
re-run the model) from checkpoint-internal validation (which does), and
do not read differences between distinct scene populations as causal
effects.

\section{Model Architecture}\label{sec:architecture}

AutoE2E did not start as a map- and route-conditioned model. The
starting point was a temporal camera-only BEV planner built on BEVFormer
v2~\cite{ref-bevformerv2}; the working group's contribution on top of
it is a navigation pathway that conditions this baseline on an HD map
and a route. This section covers both phases, the deployment contract
they share, the control planner, and the resulting parameter budget.

\begin{figure}[h]
\centering
\includegraphics[width=\linewidth]{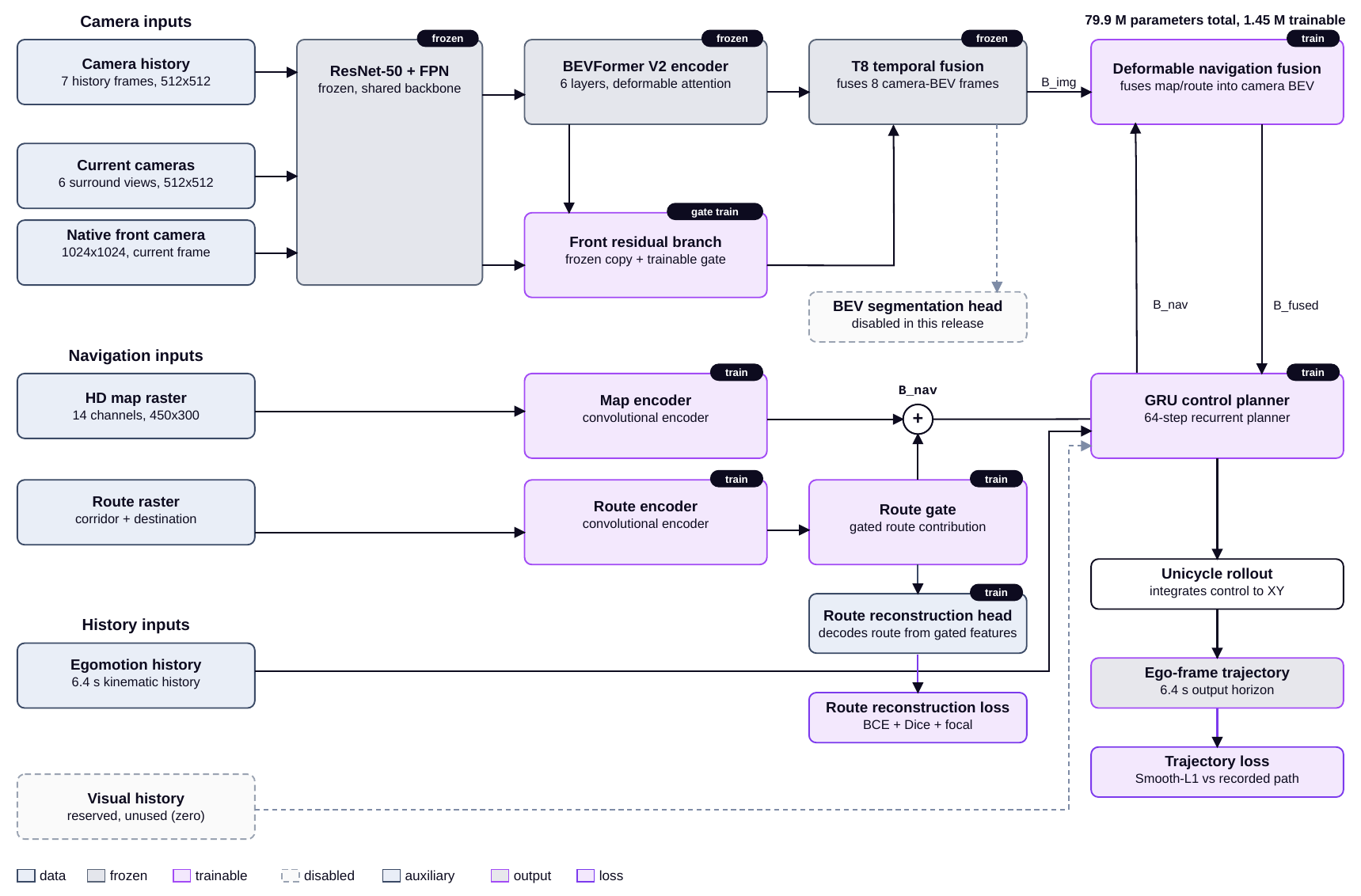}
\caption{The evaluated map- and route-conditioned AutoE2E configuration.
At each of eight times, six 512$\times$512 camera views are processed:
the current time plus seven history times. A 1024$\times$1024 front
view is added only at the current time. A frozen ResNet-50 + FPN and
six-layer single-frame BEVFormer V2 encoder are applied independently
per time, followed by T8 convolutional fusion across the eight BEVs; a
navigation pathway encodes the 14-channel map and the 2-channel route
separately, gated summation, deformable navigation fusion, 6.4\,s of
ego-motion history, a GRU control planner, and unicycle integration. All
reported counts are parameter counts measured by instantiating the
implementation (79,906,522 total; 1,448,582 trainable). Dashed boxes
denote branches present in the checkpoint but disabled.}\label{fig-arch}
\end{figure}

\subsection{Deployment Contract}\label{deployment-contract}

At every frame, the deployed AutoE2E policy receives: six-view camera
images (the current frame plus 7 history frames at 0.5\,s intervals,
each 512$\times$512), a 1024$\times$1024 front camera image for the
current frame, ego-centered HD-map rasters (14 channels) and a
selected-route raster (2 channels), and 6.4\,s of ego-motion history at
10\,Hz. The output is 6.4\,s of acceleration and curvature at 10\,Hz.
The primary evaluation is conducted under the condition in which all of
these inputs are available; because situations in which the map or the
route is missing can occur in operation, behavior from cameras alone is
measured as a separate track (Sec.~\ref{sec:benchmarks}).

\subsection{Phase 1: Temporal Camera BEV Baseline}\label{sec:baseline}

The baseline AutoE2E configuration -- the architecture the working group
started from -- is the frozen BEVFormer V2 T8 temporal camera pathway
alone, with no HD-map or route input: six camera views per frame, fused
temporally across eight timesteps, feeding directly into the control
planner (Sec.~\ref{sec:control-planner}).

The camera pathway builds on the official BEVFormer V2 R50 T8 checkpoint
\cite{ref-bevformerv2}. A four-level, 256-channel feature pyramid is
formed from the last three stages of ResNet-50~\cite{ref-resnet}, and a
six-layer BEVFormer encoder updates 300$\times$200 learned BEV queries
(256-dimensional, 15.36\,M parameters). The BEV grid covers X from
$-$60\,m to 120\,m and Y from $-$60\,m to 60\,m in ego coordinates, at
0.6\,m cells. Each layer consists of
deformable self-attention over two queues obtained by duplicating the
current BEV (8 heads, 4 points each), multi-scale spatial cross-attention
that maps reference points at four pillar heights ($z \in [-5, 3]$\,m)
into the images through a calibrated projection operator (8 heads, 4
levels, 8 points each), a 512-dimensional FFN, and three LayerNorms. The
projection operator implements pinhole and fisheye (F-theta) models, so
this evaluation is not calibration-free.

\begin{figure}[h]
\centering
\includegraphics[width=\linewidth]{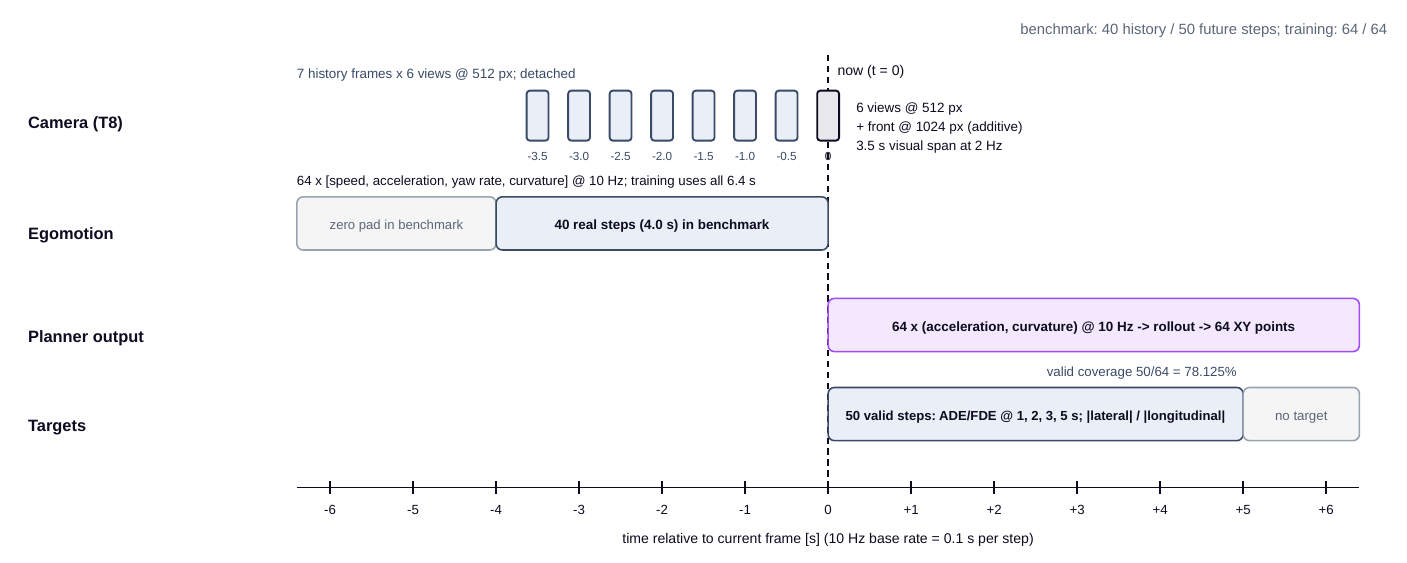}
\caption{Temporal contract. The camera branch uses 8 frames at 0.5\,s
intervals (3.5\,s), the ego-motion history uses 64 steps at 10\,Hz
(6.4\,s), and the output is 64 steps at 10\,Hz (6.4\,s). Under the
KITScenes benchmark protocol, only 40 history steps (4.0\,s, the
remainder zero-padded) and 50 future steps (5.0\,s) are real data, so
the effective horizon coverage of external replay is
50/64 = 78.125\%.}\label{fig-temporal}
\end{figure}

In the T8 temporal configuration, seven history frames at 0.5\,s
intervals ($t{-}3.5$\,s to $t{-}0.5$\,s) and the current frame are
passed through the same backbone and encoder, with gradients detached
for the history BEVs. The eight BEVs are concatenated into 2048
channels, passed through a three-block ResNet-style fusion module with
512 intermediate channels, and projected to 256 channels to obtain
$B_\text{img}$:
\begin{equation}\label{eq-temporal}
B_{img} = T\left( B_{-7},B_{-6},\ldots,B_{-1},B_{0} \right).
\end{equation}
The camera pathway therefore observes 3.5\,s at 2\,Hz, a temporal span
different from the 6.4\,s ego-motion history used by the planner
(Fig.~\ref{fig-temporal}).

The current-frame front camera is additionally processed at
1024$\times$1024: a branch consisting of a frozen copy of the last
encoder layer's cross-attention computes a content-dependent delta
$\delta$ from the high-resolution front features and adds it as
$B_0 = B_\text{enc} + \tanh(g) \odot \delta$. Since $g$ is
256-dimensional and zero-initialized, the output of the pretrained
camera pathway is unchanged at the start of training; these 256
parameters are the only camera-side parameters excluded from freezing.

\subsection{Phase 2: Map- and Route-Conditioned Extensions}
\label{sec:current-version}

The working group's contribution on top of the baseline above is a
navigation pathway that encodes an HD-map raster and a route raster
separately, and fuses them into the camera BEV through a gated residual
and deformable cross-attention. This is the current working version of
AutoE2E, and the one all results in this paper are reported for.

\begin{figure}[h]
\centering
\includegraphics[width=\linewidth]{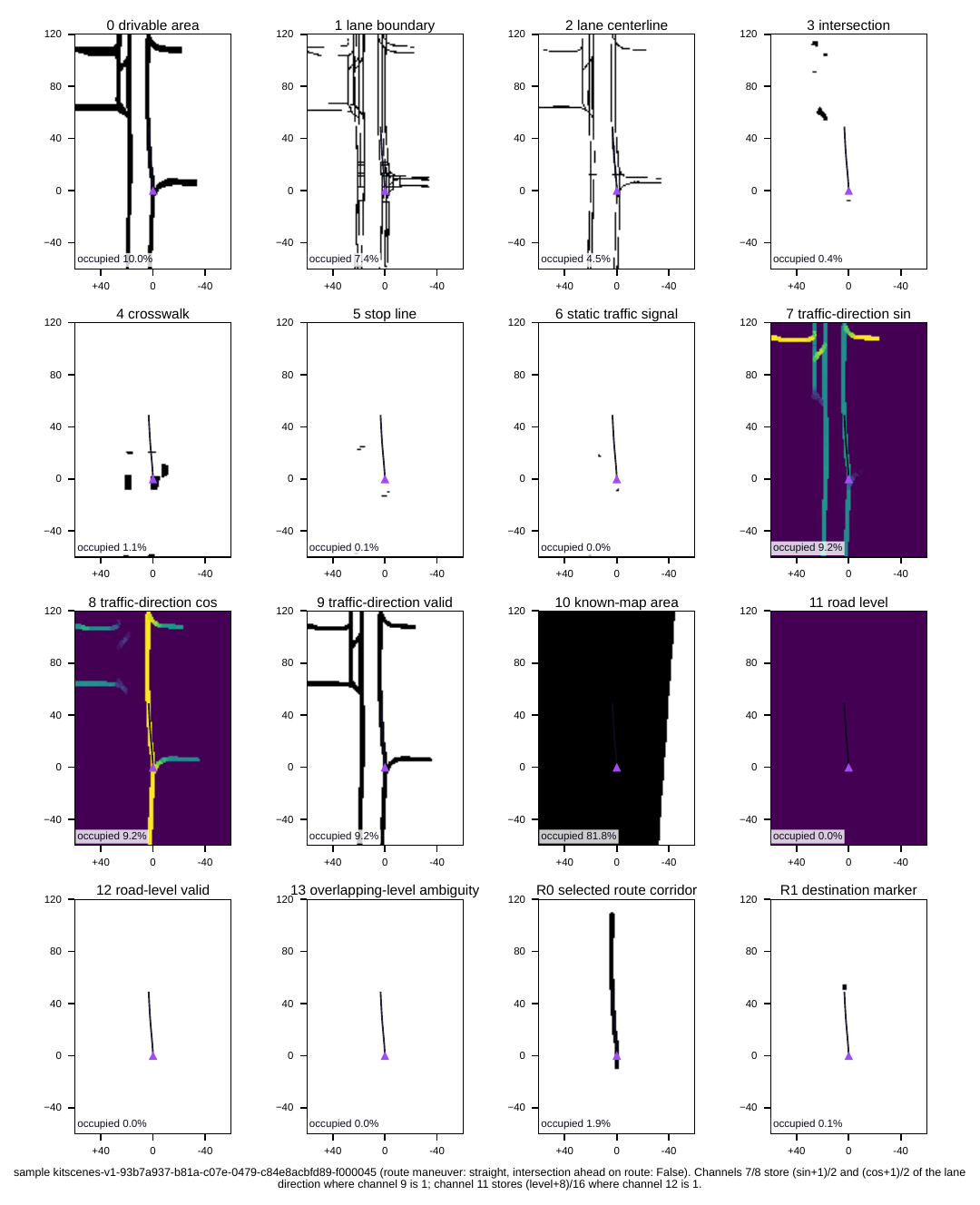}
\caption{The 14 map channels and 2 route channels for one KITScenes
validation sample. These are real data taken from the working group's
public Hugging Face dataset viewer for this mirror; violet marks the
ego position, obsidian the recorded 5\,s trajectory. Road-level channels
11--13 were empty in this sample.}\label{fig-channels}
\end{figure}

The HD map and the route are represented as ego-centric semantic
rasters, 450$\times$300 cells at 0.4\,m/px, covering the same extent as
the camera BEV. Table~\ref{tab-channels} defines the 14 map-raster and
2 route-raster channels; Fig.~\ref{fig-channels} renders them for a real
KITScenes validation sample. Binary channels store 1, direction
channels store $(\sin\theta+1)/2$ and $(\cos\theta+1)/2$, and road level
stores $(\text{clip}(\text{level}, -8, 8)+8)/16$, so all values lie in
$[0,1]$. The route corridor is drawn from the sequence of lane
centerlines with a width of 3.5\,m (truncated 10\,m behind the ego
vehicle), and the destination marker is a circle of radius 2.0\,m. Map
validity and route validity are explicit per-sample gates, and invalid
rasters are multiplied by zero.

\begingroup\scriptsize
\begin{longtable}[]{@{}
>{\raggedright\arraybackslash}p{0.03\linewidth}
>{\raggedright\arraybackslash}p{0.15\linewidth}
>{\raggedright\arraybackslash}p{0.42\linewidth}
>{\raggedright\arraybackslash}p{0.12\linewidth}
>{\raggedright\arraybackslash}p{0.12\linewidth}
@{}}
\caption{Definitions of the 14 map-raster and 2 route-raster channels.}\label{tab-channels}\\
\toprule\noalign{}
ch & Name & Source primitive (Lanelet2~\cite{ref-lanelet2}) & Value & Mean occ.\ [\%] \\
\midrule\noalign{}
\endhead
\bottomrule\noalign{}
\endlastfoot
0 & Drivable area & Lanelet polygons (excluding crosswalks) & binary & 15.55 \\
1 & Lane boundary & Left/right boundary lines (width 1.0\,m) & binary & 8.29 \\
2 & Lane centerline & Centerline (width 1.0\,m) & binary & 5.52 \\
3 & Intersection & Lanelet polygons with a turn\_direction attribute & binary & 1.46 \\
4 & Crosswalk & Polygons of subtype crosswalk & binary & 0.96 \\
5 & Stop line & Stop-line polylines (width 1.0\,m) & binary & 0.18 \\
6 & Static traffic light & \texttt{traffic\_light} regulatory elements (radius 1.0\,m) & binary & 0.07 \\
7--8 & Heading sin/cos & Centerline orientation (width 3.5\,m) & $(\sin,\cos)$ & 14.72 \\
9 & Heading valid & Same as above & binary & 14.72 \\
10 & Known map area & Map boundary polygon & binary & 96.98 \\
11--13 & Road level, valid, ambiguity & \texttt{layer}/\texttt{level} attributes & mixed & 0.00 \\
R0 & Selected route corridor & Centerlines of the route lane sequence (width 3.5\,m) & binary & 1.45 \\
R1 & Destination marker & Route endpoint (radius 2.0\,m) & binary & 0.05 \\
\end{longtable}
\endgroup

\begin{figure}[ht]
\centering
\includegraphics[width=\linewidth]{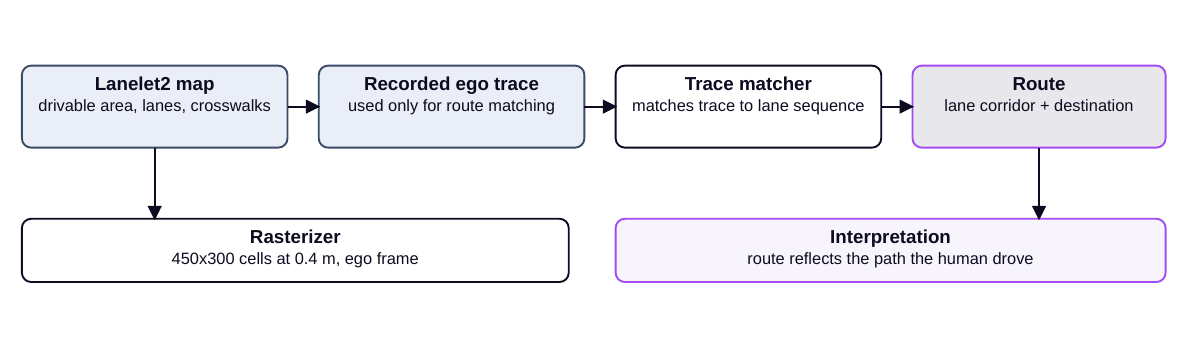}
\caption{Generation of the map and route rasters in KITScenes.
Primitives are extracted from the Lanelet2 map, and the ego trajectory
over the whole scene is fitted to a lane sequence using HMM-like
transition costs to yield the route. The route is not the output of a
live route planner but a retrospective lane-level driving
intent.}\label{fig-route}
\end{figure}

Fig.~\ref{fig-route} shows how these rasters are produced upstream of
the model. The KITScenes route is the lane sequence obtained by fitting
the ego trajectory over the whole scene to the Lanelet2 map, and the
destination is the ego position at the end of the scene. The future ego
points themselves are not drawn, but the route is a retrospectively
determined sequence of already-driven lanes -- that is, it equals the
intent the driver actually followed. Because this fact bears directly
on how route conditioning should be interpreted, we return to it in
Sec.~\ref{sec:limitations}.

\begin{figure}[h]
\centering
\includegraphics[width=\linewidth]{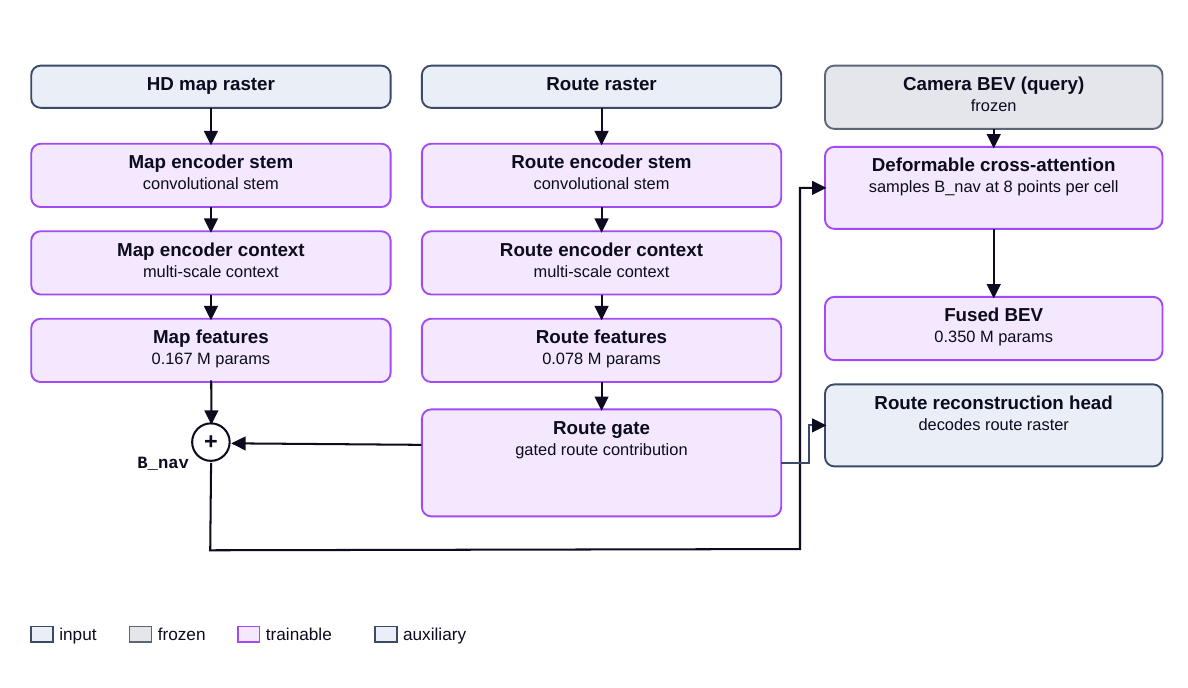}
\caption{Internal structure of the separate navigation encoding and the
deformable navigation fusion. Left: map encoder $E_\text{map}$
(0.167\,M) and route encoder $E_\text{route}$ (0.078\,M), the
per-channel gate, and the addition. Right: deformable cross-attention
that uses each cell of $B_\text{img}$ as a query and samples 8 points on
$B_\text{nav}$ (8 heads, 256 channels, 0.350\,M). The output projection
and the final FFN linear layer are zero-initialized, so the fusion is
exactly the identity at the start of training.}\label{fig-navfusion}
\end{figure}

The static map and the selected route have separate encoders
(Fig.~\ref{fig-navfusion}). The map encoder $E_\text{map}$ is a fully
convolutional network in which a 5$\times$5 convolutional stem (96
channels) and a context path reduced to strides 2 and 4 by depthwise
convolutions (192 channels) are aligned to 256 channels by 1$\times$1
projections and added, with the input bilinearly resampled to
300$\times$200. The route encoder $E_\text{route}$ is a lightweight
convolutional network with 96 hidden channels, designed so the thin
route corridor and destination features are not lost to coarse pooling.
The route contribution is controlled by a 256-dimensional per-channel
gate and added to the encoded map:
\begin{equation}\label{eq-nav}
B_{nav} = E_{map}(M) + \sigma(g) \odot E_{route}(R).
\end{equation}
The gate $g$ is zero-initialized, so its weight at the start of
training is 0.5.

The navigation BEV is fused into the camera BEV by deformable
cross-attention~\cite{ref-deformable}:
\begin{equation}\label{eq-fuse}
B_{fused} = B_{img} + F_{deform}\left( B_{img},B_{nav} \right).
\end{equation}
Each image-BEV cell acts as a query, predicts 8 offsets from its own
position as the reference point, bilinearly samples $B_\text{nav}$, and
aggregates the samples with 8-head softmax weights -- reducing what
would be $O(N^2)$ for dense cross-attention over 60,000 cells to
$O(NK)$. Because the output projection and the final FFN linear layer
are zero-initialized, the fusion does not alter the representation of
the pretrained camera pathway at the start of training, acting as a
stabilization mechanism when attaching new navigation features to a
frozen visual BEV representation.

\subsection{Control Planner}\label{sec:control-planner}

\begin{figure}[h]
\centering
\includegraphics[width=0.85\linewidth]{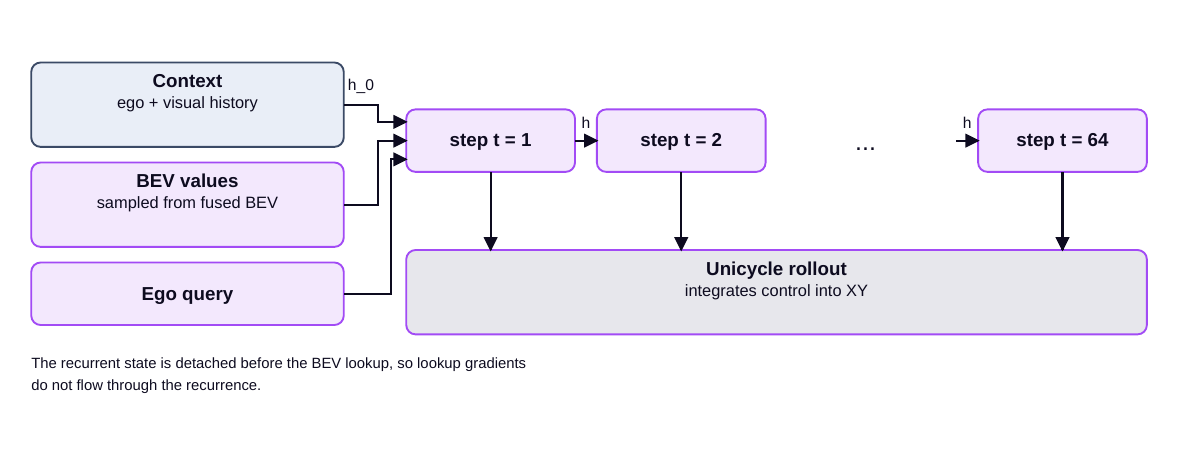}
\caption{The deterministic GRU control planner (0.835\,M parameters).
The Bezier and flow-matching planner heads present in the codebase are
not used by the evaluated checkpoints.}\label{fig-planner}
\end{figure}

Both phases feed the same fused BEV into one control planner. The
ego-motion history $h_\text{ego}$ is the flattened 256-dimensional
vector of $[$speed, acceleration, yaw rate, curvature$]$ over 64 steps
at 10\,Hz. The model API retains an 896-dimensional visual-history input
$h_\text{vis}$, reserved for the future world-action band
(Sec.~\ref{sec:future}); the evaluated checkpoints feed it as zero. The
planner is a deterministic GRU (Fig.~\ref{fig-planner}):
\begin{align}
c_{0} &= P_{ego}(h_{ego}) + P_{vis}(h_{vis}), &
a_{t} &= A_{deform}\left( q_{t},B_{fused} \right),\; q_{t} = \mathrm{sg}(h_{t-1}) + e, \\
h_{t} &= \mathrm{GRU}\left( a_{t},h_{t-1} \right), &
u_{t} &= W_{u}h_{t} = \left( a_{t}^{acc},\kappa_{t} \right).
\end{align}
At each of 64 future steps, 16 points on the fused BEV are attended to
deformably and the GRU updates its state and emits acceleration and
curvature. A semi-implicit unicycle model then integrates the control
sequence into ego-frame XY ($\Delta t = 0.1$\,s):
\begin{equation}\label{eq-roll}
v_{t} = \max\left( v_{t-1} + a_{t}^{acc}\Delta t,\,0 \right),\quad
\theta_{t} = \sum_{i\le t} v_i \kappa_i \Delta t,\quad
x_{t} = \sum_{i\le t} v_i \cos\theta_i \Delta t,\quad
y_{t} = \sum_{i\le t} v_i \sin\theta_i \Delta t.
\end{equation}

The route-reconstruction head decodes two channels of logits at
450$\times$300 from the gated route contribution, supervised by the
mean of BCE and soft Dice for the corridor and a heatmap focal loss
(weight 0.25) for the destination. The trajectory loss is Smooth-L1
($\beta = 1$\,m) between the integrated XY and the recorded trajectory.
All reported checkpoints were trained with a trajectory weight of 1.0,
a route-reconstruction weight of 1.0, and a BEV-segmentation weight of
0.0 -- the segmentation head exists in the checkpoint but is not jointly
optimized in this release.

\subsection{Model Size}\label{sec:size}

\begin{figure}[h]
\centering
\includegraphics[width=0.95\linewidth]{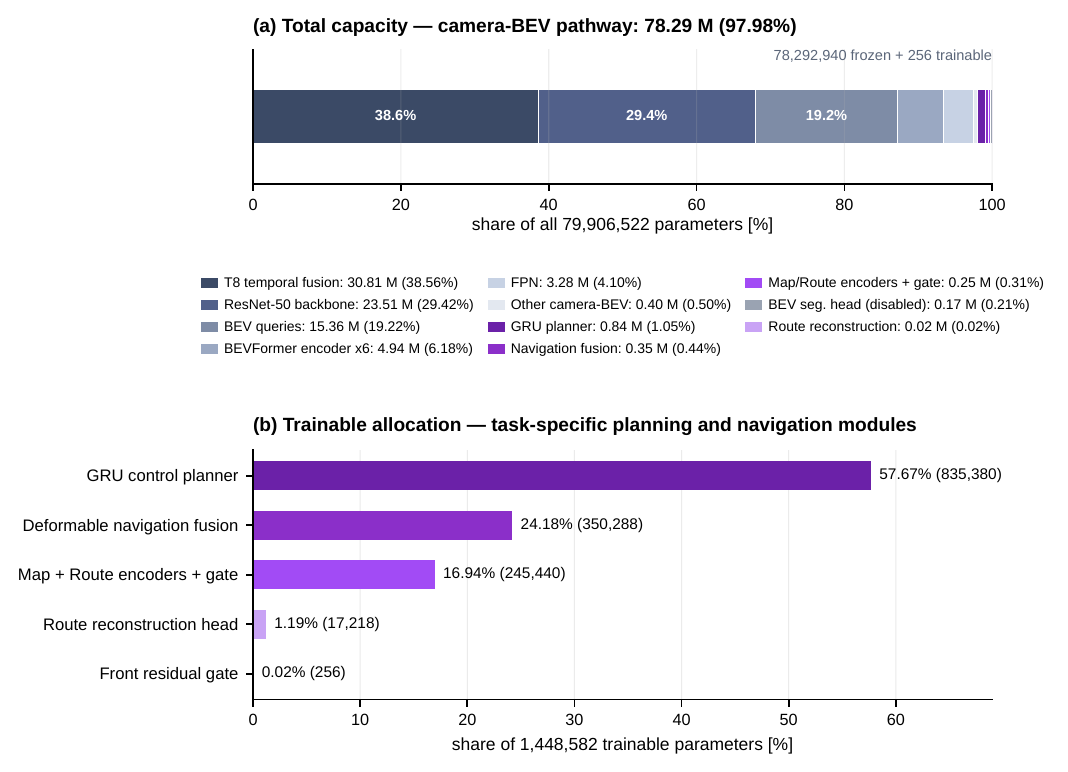}
\caption{Parameter allocation. (a) The frozen camera-BEV pathway is
97.98\% of all parameters. (b) Among the 1.45\,M trainable parameters,
the GRU planner is 57.7\%, deformable navigation fusion 24.2\%, and the
map/route encoders plus route gate 16.9\%.}\label{fig-params}
\end{figure}

Table~\ref{tab-params} and Fig.~\ref{fig-params} report parameter
counts obtained by instantiating the model in the evaluated six-view
configuration. In total-capacity terms, the model is almost entirely a
pretrained, frozen visual representation: 97.98\% of all parameters
(78,292,940 of 79,906,522) belong to the frozen camera-BEV pathway,
split mainly between T8 temporal fusion, ResNet-50, and the learned BEV
queries. The remaining 1,448,582 parameters (1.81\%) are updated during
trajectory training, allocated across the GRU planner, the deformable
navigation fusion, the map/route encoders, and the route-reconstruction
head. At 79.9\,M total parameters, AutoE2E is roughly two orders of
magnitude smaller than the $\sim$4\,B-parameter and $\sim$10\,B-parameter
vision-language-action planners it is benchmarked against
(Sec.~\ref{sec:benchmarks}) -- a deliberately light footprint for
real-time, on-vehicle inference rather than research-scale capability.

\begingroup\scriptsize
\begin{longtable}[]{@{}lll@{}}
\caption{Parameter counts for the evaluated six-view configuration.}\label{tab-params}\\
\toprule\noalign{}
Component & Parameters & State \\
\midrule\noalign{}
\endfirsthead
\toprule\noalign{}
Component & Parameters & State \\
\midrule\noalign{}
\endhead
\bottomrule\noalign{}
\endlastfoot
ResNet-50 backbone & 23,508,032 & frozen \\
Feature pyramid (4 levels) & 3,278,592 & frozen \\
BEV queries 300$\times$200$\times$256 & 15,360,000 & frozen \\
Row, column, level, and camera embeddings & 66,560 & frozen \\
Encoder, 6 layers & 4,940,928 & frozen \\
Front cross-attention (copy) & 328,960 & frozen \\
Front residual gate & 256 & trained \\
T8 temporal fusion & 30,809,856 & frozen \\
Map encoder $E_\text{map}$ & 167,200 & trained \\
Route encoder $E_\text{route}$ & 77,984 & trained \\
Route gate & 256 & trained \\
Deformable navigation fusion & 350,288 & trained \\
GRU control planner & 835,380 & trained \\
Route reconstruction head & 17,218 & trained \\
BEV segmentation head (weight 0.0) & 165,000 & frozen \\
Total / trainable & 79,906,522 / 1,448,582 & --- \\
\end{longtable}
\endgroup

\section{Training}\label{sec:training}

\subsection{Datasets}\label{sec:datasets}

Two datasets anchor training and evaluation. \textbf{nuPlan}~\cite{ref-nuplan},
a large-scale, multi-city closed-loop planning benchmark, is used for
the initial trajectory-and-route training stage with the camera BEV
backbone frozen at its official pretrained weights. \textbf{KITScenes}~\cite{ref-kitscenes},
a multimodal European driving dataset from the Karlsruhe Institute of
Technology (KIT) pairing high-fidelity camera and sensor data with
complete Lanelet2~\cite{ref-lanelet2} HD maps, is used to fine-tune the
nuPlan-trained checkpoint to the Karlsruhe road network and camera
configuration, and for the primary external evaluation
(Sec.~\ref{sec:benchmarks}). Both datasets carry their own licenses,
which the released checkpoint weights inherit; the model architecture,
training code, and evaluation pipeline are Apache-2.0 regardless of
which dataset trained a given checkpoint.

The split manifest identifies the KITScenes fine-tuning source as the
official training split: 129 empty scenes are excluded from 533
available train scenes, leaving 404 eligible scenes and 42,667 samples.
A scene-level 90\%/10\% split yields 38,847 samples from 364 scenes for
training and 3,820 samples from 40 scenes for frozen internal
validation. Because these 404 scenes belong to the official training
split, their scene IDs are disjoint from the official 117-scene Val and
206-scene Test populations used for evaluation.

\subsection{How It Was Trained}\label{sec:how-trained}

\begin{figure}[h]
\centering
\includegraphics[width=\linewidth]{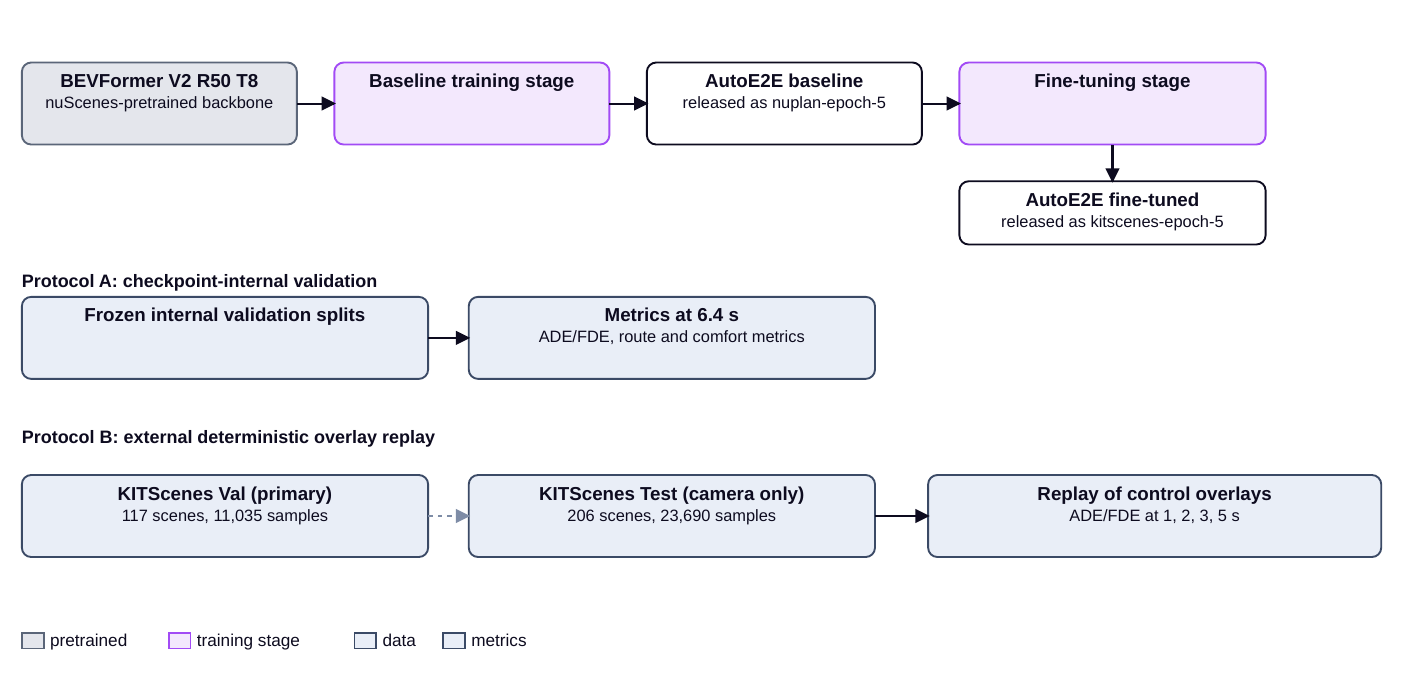}
\caption{Checkpoint lineage (top) and the two evaluation protocols
(bottom). Protocol A is checkpoint-internal validation that runs the
model; Protocol B is replay of the released deterministic control
overlays. The two are never mixed in a single ranking table. Val and
Test are different scene populations.}\label{fig-protocol}
\end{figure}

Training proceeds in two stages (Fig.~\ref{fig-protocol}). The
\textbf{baseline} stage trains the trajectory and route pathways on
nuPlan with the camera BEV frozen at its official pretrained weights:
8 distributed GPU workers, seed 149, learning rate \texttt{1e-4},
trajectory weight 1.0, route-reconstruction weight 1.0, BEV-segmentation
weight 0.0, and 1,024 internal validation samples. The resulting
checkpoint is released as \texttt{nuplan-epoch-5}. The
\textbf{fine-tuning} stage then freezes that checkpoint as parent and
continues training on KITScenes at learning rate \texttt{3e-5}, with
the same 8-worker, seed-149, loss-weight configuration, over the
38,847-sample / 364-scene training split described above, validated
internally on 3,820 samples / 40 scenes in BF16 precision. The result is
released as \texttt{kitscenes-epoch-5}, the checkpoint fine-tuned for
the Karlsruhe road network and camera configuration, and the one
reported as the primary result throughout Sec.~\ref{sec:benchmarks}. An
optional BEV-only training stage (unfrozen camera BEV, BEV-segmentation
loss) exists in the code path; its use as parent of the released
baseline checkpoint is not recorded in the available metadata. Neither
released checkpoint passed the repository's automated checkpoint
quality gate.

\subsection{Where It Was Trained}\label{sec:infra}

Training runs on an EKS Auto Mode Kubernetes cluster on AWS, chosen
over a standard EKS node group plus Karpenter setup for its built-in
CNI, autoscaler, and load-balancer controller, which removes most of
the day-two operational overhead of running GPU infrastructure.
Imitation learning uses Ray Train jobs admitted by Kueue's GPU-quota
scheduler onto \texttt{p5en.48xlarge} Capacity Block reservations (4- or
8-GPU jobs, whole-GPU allocation to preserve intra-instance
communication), with a smaller retained \texttt{g6.2xlarge} pool
reserved for inference, validation, and canary jobs. Closed-loop
simulation was evaluated and deliberately not adopted for this release:
CARLA's Vulkan dependency is incompatible with the Bottlerocket AMI that
EKS Auto Mode's GPU nodes run, so the working group's closed-loop
\emph{training} path instead goes through offline reinforcement learning
(Implicit Q-Learning) on the same WebDataset shards used for imitation
learning, requiring no simulator. Closed-loop \emph{evaluation} is
pursued separately, through NVIDIA's AlpaSim; its first result, a
diagnosed closed-loop failure mode rather than a passing validation, is
reported in Sec.~\ref{sec:future}. The likely launch configuration for training
the released checkpoints, traced from the training branch, is a single
\texttt{p5en.48xlarge} ($8\times$H200) or \texttt{p5.48xlarge}
($8\times$H100) Capacity Block, 8-way distributed, consistent with the
epoch, learning-rate, and seed values above; this is an inferred
hardware configuration, not a recorded one, and total training
wall-clock time and GPU-hours are not available in the evidence used
for this paper. Every stage -- data ingest, imitation-learning training,
open-loop evaluation gating, and offline-RL refinement -- logs
experiments, metrics, and artifacts to MLflow, orchestrated as Flyte
workflows; both are deployed on the same cluster and reachable through
internal, Cognito-authenticated CloudFront endpoints rather than the
public internet.

\section{The DataModelConsole Deliverable}\label{sec:dmc}

Alongside the model weights and training code, this release ships
\textbf{DataModelConsole}, live at
\url{https://d2itskdqq39tx1.cloudfront.net/}, a read-only dashboard that
replays
synchronized camera scenes with ground-truth and predicted trajectories
in both camera and bird's-eye views, next to dataset versions, MLflow
experiment runs, and Flyte pipeline executions. It is the artifact
behind both quantitative and qualitative results in this paper: the
KITScenes Val/Test benchmark tables of Sec.~\ref{sec:benchmarks} and the
qualitative examples of Sec.~\ref{sec:qualitative} (Fig.~\ref{fig-qual})
are scored and rendered from the same dataset shards and
control-overlay blobs the dashboard serves. The dashboard is
implemented as a Go API over MLflow/Flyte/S3 state and
a Next.js web frontend, and covers dataset and shard browsing, per-sample
trajectory overlays, an occupancy view, reasoning-label review (for the
future reasoning band), and experiment/run browsing -- the working
group's own tool for auditing what a checkpoint actually predicts on a
given scene before trusting a benchmark number. It is released alongside
the code at the tag in the header of this paper, and is intended as
public-facing infrastructure the broader community -- including the
institutions in Sec.~\ref{sec:collaboration} -- can use to independently
inspect AutoE2E's behavior rather than take its benchmark table on
faith.

\begin{figure}[h]
\centering
\includegraphics[width=\linewidth]{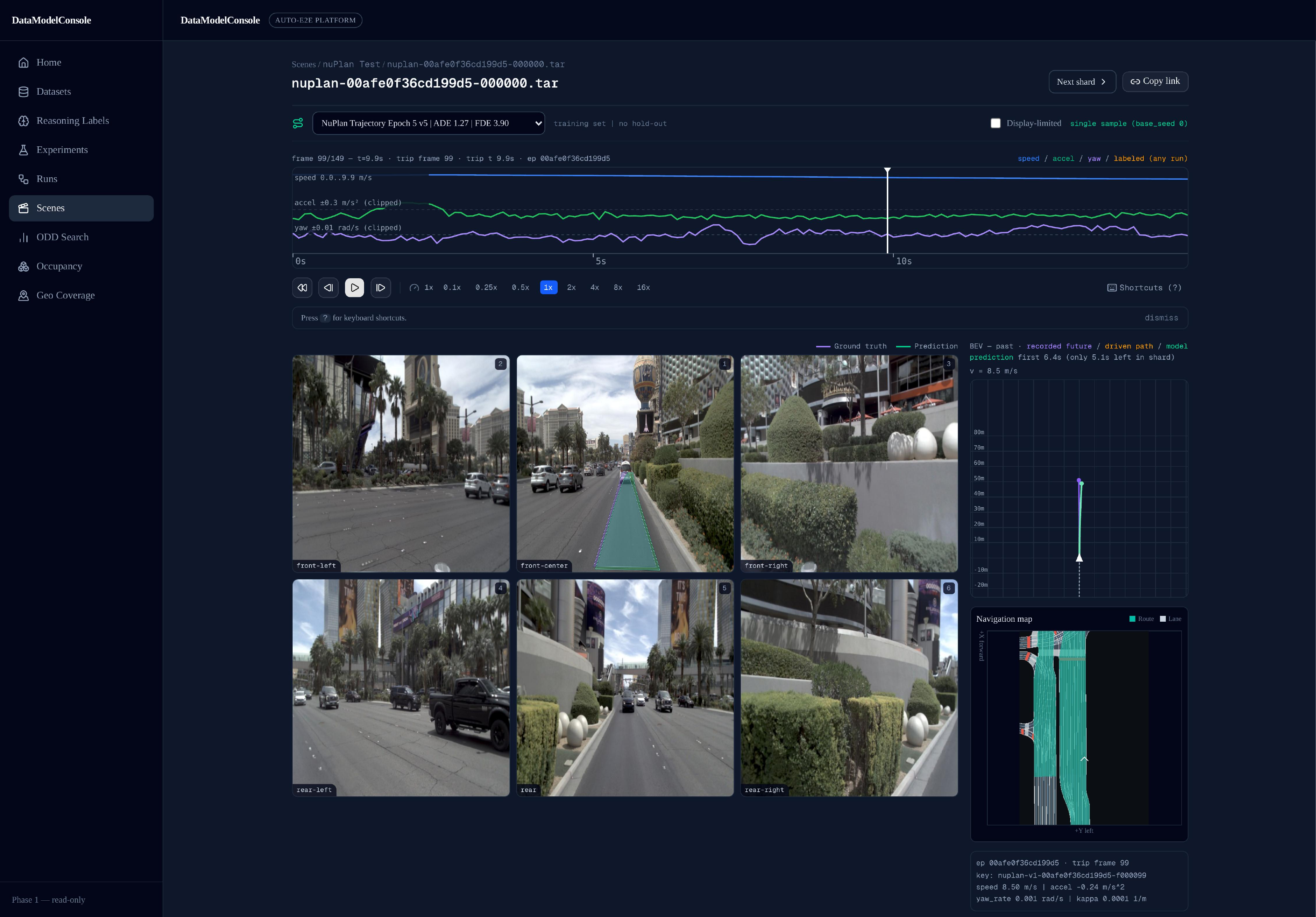}
\caption{The DataModelConsole dashboard: the six-camera mosaic, a
speed/acceleration/yaw timeline scrubber, the bird's-eye-view panel
(ground truth vs.\ prediction), and the navigation map, alongside the
sidebar's dataset, experiment, run, and occupancy
browsers.}\label{fig-dmc-screenshot}
\end{figure}

\section{Benchmarks}\label{sec:benchmarks}

\subsection{Population and Protocol}\label{sec:population-protocol}

The primary benchmark is the KITScenes validation split, where Camera +
HD Map + Route are all available, computed over the 117 scenes and
11,035 samples that satisfy strict identity with the official Map and
Route evaluation. Because the KITScenes test split does not release
maps, it is evaluated as a camera-only robustness and transfer track
over all 206 scenes and 23,690 samples. Since Val and Test are
different scene populations, the Test-versus-Val difference is not
interpreted as a causal effect of the map and route. The external
evaluation is a replay of the published deterministic control overlays;
it does not re-run the model and contains no route counterfactual
inputs, input gradients, or route-reconstruction outputs. Checkpoint
internal validation (Protocol A, Fig.~\ref{fig-protocol}) is a separate
protocol and is not mixed into the same ranking table as the external
replay (Protocol B).

\subsection{Models Used for Comparison}\label{sec:bench-models}

Table~\ref{tab-positioning} positions AutoE2E against the systems
introduced in Sec.~\ref{related-work} on a single scope comparison --
what each system takes as input and how it is evaluated, not relative
performance. Beyond AutoE2E's own two checkpoints, this paper compares
against six other driving policies evaluated on KITScenes-derived
populations: \textbf{Qwen-Drive-1.0}~\cite{ref-qwendrive}, a
vision-language-action model unifying 3D perception, driving VQA, and
motion planning, reported at approximately 4\,B parameters;
\textbf{Alpamayo\,1.5}, NVIDIA's reasoning-focused distillation-teacher
model, reported at approximately 10\,B parameters -- no public paper
describes version 1.5 specifically; \cite{ref-alpamayo} is its
predecessor, Alpamayo-R1 (also called Alpamayo~1), cited here for
background only, not as the source of the 1.5 numbers below;
\textbf{SimForge} (SIMFORGE-D1.5, a distilled Alpamayo\,1.5), evaluated
in its 7Cam-VT configuration: 7 camera slots, two of which are virtual
telephoto crops approximating cameras KITScenes does not natively
provide; \textbf{METEOR}, a model released by TIER IV as code and
weights without an accompanying paper; \textbf{Drive-JEPA}, a
video-latent (not BEV) model whose own published benchmarks are NAVSIM
and Bench2Drive rather than KITScenes or nuPlan, so its numbers here
come from the same cross-model protocol as the others rather than from
its original paper; and \textbf{UniAD}~\cite{ref-uniad}, the
planning-oriented BEV model that unifies detection, tracking, mapping,
motion forecasting, and occupancy with planning through task
interaction, evaluated here zero-shot on KITScenes since its own
published benchmark is nuScenes.

\begingroup\scriptsize
\begin{longtable}[]{@{}
>{\raggedright\arraybackslash}p{0.12\linewidth}
>{\raggedright\arraybackslash}p{0.095\linewidth}
>{\raggedright\arraybackslash}p{0.11\linewidth}
>{\raggedright\arraybackslash}p{0.11\linewidth}
>{\raggedright\arraybackslash}p{0.07\linewidth}
>{\raggedright\arraybackslash}p{0.11\linewidth}
>{\raggedright\arraybackslash}p{0.095\linewidth}
>{\raggedright\arraybackslash}p{0.125\linewidth}
@{}}
\caption{Positioning relative to representative E2E driving systems. This table delineates the scope of the claims made here and does not indicate relative performance.}\label{tab-positioning}\\
\toprule\noalign{}
System & Camera-only prim.\ eval.\ & HD map input & Route input & Temp.\ BEV & Unified det./track./occ.\ & Closed-loop eval.\ & Evaluation data \\
\midrule\noalign{}
\endhead
\bottomrule\noalign{}
\endlastfoot
UniAD~\cite{ref-uniad} & $\circ$ & --- (online map est.) & Navigation command & $\circ$ & $\circ$ & --- & nuScenes \\
VAD~\cite{ref-vad} & $\circ$ & --- (vector map est.) & Navigation command & $\circ$ & Partial & $\circ$ (CARLA) & nuScenes / CARLA \\
TransFuser~\cite{ref-transfuser} & --- (camera + LiDAR) & --- & Target point & --- & Aux.\ tasks & $\circ$ (CARLA) & CARLA \\
Qwen-Drive-1.0~\cite{ref-qwendrive} & $\circ$ & --- (online est.) & n/a & --- & $\circ$ (3D perc.) & --- & nuScenes \\
Alpamayo\,1.5 & --- (camera + nav.) & n/a & $\circ$ (nav.\ command) & n/a & --- & $\circ$ (AlpaSim) & PhysicalAI-AV, AlpaSim \\
METEOR & n/a & n/a & n/a & n/a & n/a & n/a & internal, unpublished \\
Drive-JEPA & $\circ$ & n/a & n/a & --- (video latent) & --- & $\circ$ (Bench2Drive) & NAVSIM / Bench2Drive \\
AutoE2E (this paper) & --- (secondary track) & $\circ$ (14-ch raster) & $\circ$ (2-ch raster) & $\circ$ (T8, 3.5\,s) & --- (BEV seg.\ disabled) & --- & KITScenes Val/Test, nuPlan \\
\end{longtable}
\endgroup

\subsection{Primary Evaluation: KITScenes Val}\label{sec:results-val}

\begin{figure}[h]
\centering
\includegraphics[width=\linewidth]{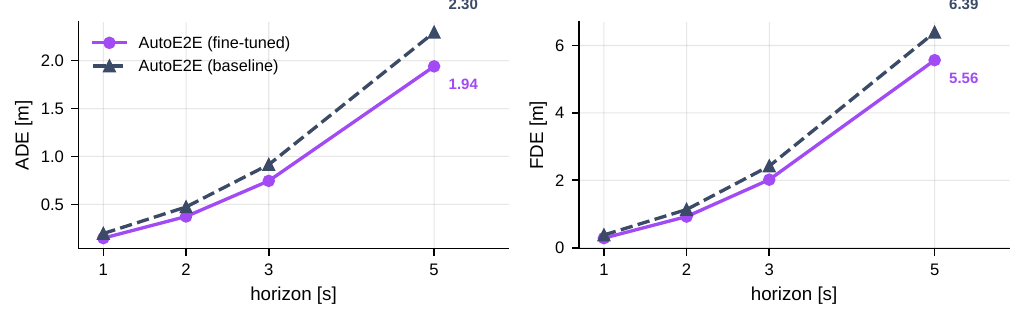}
\caption{ADE (left) and FDE (right) across horizons on KITScenes Val
(Camera + HD Map + Route; n = 11,035). Solid is the fine-tuned
checkpoint, dashed the baseline. 5\,s values are
annotated.}\label{fig-val}
\end{figure}

Table~\ref{tab-val-ade} and Fig.~\ref{fig-val} report ADE/FDE for
AutoE2E's two released checkpoints -- fine-tuned and baseline -- on
11,035 samples. The fine-tuned checkpoint is lowest at every horizon
(5\,s: ADE 1.94\,m, FDE 5.56\,m) and is the checkpoint submitted for
official KITScenes leaderboard evaluation; the baseline is inferior to
it throughout, at 5\,s FDE 6.39\,m.

\begingroup\scriptsize
\begin{longtable}[]{@{}l|llll|llll@{}}
\caption{KITScenes Val, Camera + HD Map + Route, external overlay replay (n = 11,035 samples per model). Units m; lower is better.}\label{tab-val-ade}\\
\toprule\noalign{}
Model & ADE 1\,s & ADE 2\,s & ADE 3\,s & ADE 5\,s & FDE 1\,s & FDE 2\,s & FDE 3\,s & FDE 5\,s \\
\midrule\noalign{}
\endhead
\bottomrule\noalign{}
\endlastfoot
AutoE2E (fine-tuned) & 0.1472 & 0.3729 & 0.7449 & 1.9405 & 0.2850 & 0.9239 & 2.0196 & 5.5645 \\
AutoE2E (baseline) & 0.1951 & 0.4723 & 0.9147 & 2.2951 & 0.3779 & 1.1331 & 2.4226 & 6.3885 \\
\end{longtable}
\endgroup

\subsection{Secondary Evaluation: KITScenes Test (Camera Only)}

\begin{figure}[h]
\centering
\includegraphics[width=\linewidth]{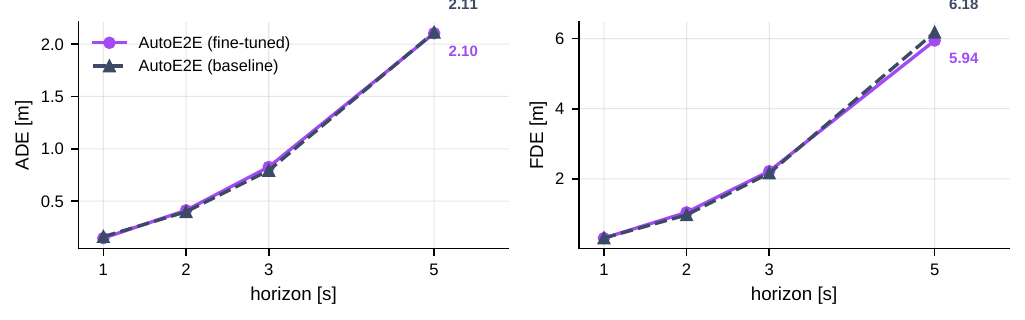}
\caption{ADE and FDE on KITScenes Test (Camera only; n = 23,690). This
is a different scene population from Val, so differences with respect
to Fig.~\ref{fig-val} do not imply a causal effect of map and
route.}\label{fig-test}
\end{figure}

Table~\ref{tab-test-ade} and Fig.~\ref{fig-test} report results on the
camera-only track (23,690 samples) for the same two checkpoints. The
fine-tuned checkpoint has the lowest reported means at 1\,s and 5\,s,
whereas the baseline is lowest at 2\,s and 3\,s. As throughout, Val and
Test are different scene populations, so this table is not a map/route
ablation against Table~\ref{tab-val-ade}. The fine-tuned checkpoint was
trained under a condition in which the map and the route are provided,
so it is conceivable that it becomes relatively brittle when they are
absent; because Test and Val contain different scenes, geographies, and
motion distributions, this gap cannot be quantified as the effect of
missing map and route inputs.

\begingroup\scriptsize
\begin{longtable}[]{@{}l|llll|llll@{}}
\caption{KITScenes Test, Camera only (map and route unavailable by construction), external overlay replay (n = 23,690 samples per model).}\label{tab-test-ade}\\
\toprule\noalign{}
Model & ADE 1\,s & ADE 2\,s & ADE 3\,s & ADE 5\,s & FDE 1\,s & FDE 2\,s & FDE 3\,s & FDE 5\,s \\
\midrule\noalign{}
\endhead
\bottomrule\noalign{}
\endlastfoot
AutoE2E (fine-tuned) & 0.1463 & 0.4121 & 0.8263 & 2.1042 & 0.3107 & 1.0453 & 2.2199 & 5.9422 \\
AutoE2E (baseline) & 0.1602 & 0.3979 & 0.7924 & 2.1120 & 0.3134 & 0.9727 & 2.1621 & 6.1751 \\
\end{longtable}
\endgroup

\subsection{Cross-Model Benchmark}\label{sec:cross-model}

\begin{figure}[h]
\centering
\includegraphics[width=0.92\linewidth]{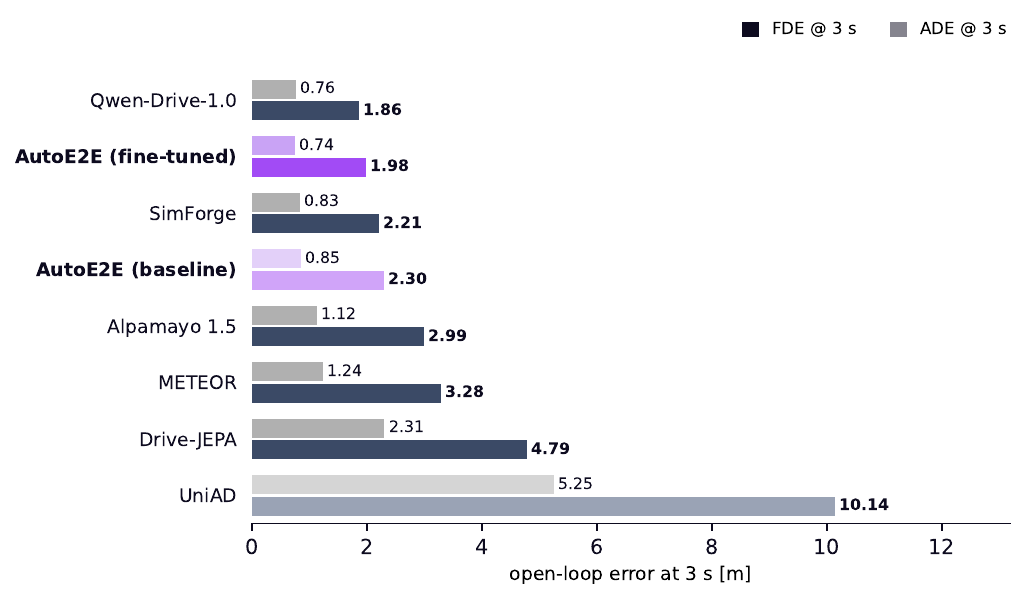}
\caption{ADE/FDE at the 3\,s horizon for the comparison in
Table~\ref{tab-crossmodel}, sorted by FDE@3\,s, for reference only, all
on the 202-keyframe population. UniAD, marked~\textsuperscript{\dag}
and shown desaturated, is evaluated zero-shot and cross-domain with no
KITScenes-specific retraining, so its row is not part of a controlled
comparison against the other rows; see Table~\ref{tab-crossmodel} for
the full-horizon comparison.}\label{fig-crossmodel}
\end{figure}

Table~\ref{tab-crossmodel} and Fig.~\ref{fig-crossmodel} compare AutoE2E
against the six other driving policies of Sec.~\ref{sec:bench-models},
sorted into a single list by FDE@3\,s. All six -- Qwen-Drive-1.0,
SimForge, Alpamayo\,1.5, METEOR, Drive-JEPA, and UniAD -- together with
both of AutoE2E's own checkpoints, are scored on the same
non-overlapping population of 202 anchor keyframes built from the 117
Val episodes, against the same canonical KITScenes ego-pose ground
truth; AutoE2E's baseline checkpoint was moved onto this population for
this release, so for the first time it is directly comparable to the
other six. A separate evaluation on the full 11,035-sample Val
population (Table~\ref{tab-val-ade}) was also run to characterize
AutoE2E's behavior over that longer, non-anchor-sampled population;
those results are not repeated in this cross-model comparison, which is
restricted throughout to the 202-keyframe population above. 

\newpage 

UniAD,
marked~\textsuperscript{\dag}, shares the 202-keyframe population but is
evaluated zero-shot, with no KITScenes-specific retraining, against a
model whose own training and benchmark is nuScenes; its position
likewise should not be read as a controlled comparison. Camera count and
preprocessing also differ by model -- AutoE2E uses six views at a fixed
square resolution, Qwen-Drive-1.0 used three front cameras,
Alpamayo\,1.5 used four, and METEOR a single frame with no map or route
input at all. Among all eight models scored on the 202-keyframe
population, AutoE2E's fine-tuned checkpoint reports the lowest ADE and
FDE at 1\,s and 2\,s, and the lowest ADE at 3\,s; Qwen-Drive-1.0 reports
the lowest FDE at 3\,s and the lowest ADE and FDE at 5\,s, with AutoE2E's
fine-tuned checkpoint second among the models that report 5\,s -- at
roughly two orders of magnitude fewer parameters than Qwen-Drive-1.0.
AutoE2E's zero-shot baseline checkpoint ranks fourth among the eight by
FDE@3\,s, behind Qwen-Drive-1.0, its own fine-tuned checkpoint, and
SimForge. UniAD's errors are several times higher than the rest of the
202-keyframe population at every horizon it reports; given the
zero-shot, cross-domain caveat above, this is read as a plausible
domain-shift effect rather than a controlled capability comparison, and
for this reason UniAD is shown desaturated in Fig.~\ref{fig-crossmodel}
rather than omitted. METEOR, Drive-JEPA, and UniAD do not report a 5\,s figure.
Hold-last-action on the 202-keyframe population is not yet computed in
the evidence available for this paper. A constant-velocity reference on
this same population is available from the UniAD benchmark package
(ADE/FDE 0.168/0.406\,m at 1\,s, 0.557/1.484\,m at 2\,s, and
1.162/3.195\,m at 3\,s); Sec.~\ref{sec:tum} separately reports a
constant-velocity reference on TUM's EDGAR data. Full
methodology for every row -- inputs, ground truth, population, and
whether the number is independently reproducible or
contributor-reported -- is documented per model in the working group's
repository under \texttt{Model/accuracy\_benchmark/}, with the
exception of Drive-JEPA's 202-keyframe result, which at the time of
writing is tracked in the working group's issue tracker (issue
\#241) pending being folded into its benchmark package alongside the
other models.

\begingroup\scriptsize
\begin{longtable}[]{@{}
>{\raggedright\arraybackslash}p{0.15\linewidth}
>{\raggedright\arraybackslash}p{0.18\linewidth}
>{\raggedright\arraybackslash}p{0.08\linewidth}
>{\raggedright\arraybackslash}p{0.135\linewidth}
>{\raggedright\arraybackslash}p{0.135\linewidth}
>{\raggedright\arraybackslash}p{0.135\linewidth}
>{\raggedright\arraybackslash}p{0.135\linewidth}
@{}}
\caption{Open-loop ADE/FDE (m) across eight driving policies (AutoE2E
counted as two checkpoints), sorted by FDE@3s, all scored on the
202-anchor non-overlapping keyframe population (``202 kf''), against
canonical KITScenes ego-pose ground truth. AutoE2E and UniAD were each
also evaluated on the full 11,035-sample dense population (Table~\ref{tab-val-ade}
for AutoE2E; \texttt{Model/accuracy\_benchmark/uniad/} for UniAD); those
results are not repeated here.
\textsuperscript{\dag} marks UniAD, evaluated zero-shot and cross-domain
with no KITScenes-specific retraining; its row is not part of a
controlled comparison against the unmarked rows, despite sharing this
sorted list. ``---'' marks a horizon the source did not
report.}\label{tab-crossmodel}\\
\toprule\noalign{}
Model & Setting & Evaluated on & ADE/FDE@1s & ADE/FDE@2s & ADE/FDE@3s & ADE/FDE@5s \\
\midrule\noalign{}
\endfirsthead
\toprule\noalign{}
Model & Setting & Evaluated on & ADE/FDE@1s & ADE/FDE@2s & ADE/FDE@3s & ADE/FDE@5s \\
\midrule\noalign{}
\endhead
\bottomrule\noalign{}
\endlastfoot
Qwen-Drive-1.0~\cite{ref-qwendrive} & first of 6 trajectories & 202 kf & 0.173/0.349 & 0.422/0.981 & 0.764/1.862 & 1.685/4.331 \\
\textbf{AutoE2E (fine-tuned)} & single & 202 kf & \textbf{0.147/0.296} & \textbf{0.380/0.933} & \textbf{0.744/1.980} & \textbf{1.883/5.477} \\
SimForge (7Cam-VT) & first of 6 & 202 kf & 0.159/0.336 & 0.427/1.047 & 0.831/2.209 & 2.077/5.756 \\
\textbf{AutoE2E (baseline)} & single, zero-shot & 202 kf & \textbf{0.176/0.339} & \textbf{0.431/1.055} & \textbf{0.854/2.302} & \textbf{2.232/6.318} \\
Alpamayo\,1.5 & first of 6 trajectories & 202 kf & 0.187/0.409 & 0.551/1.441 & 1.124/2.988 & 2.564/6.307 \\
METEOR & top-1 of 3 paths & 202 kf & 0.219/0.484 & 0.630/1.574 & 1.240/3.282 & --- \\
Drive-JEPA & top-1 of 32, perception-based v2, 20\,m & 202 kf & 0.745/1.402 & 1.502/2.965 & 2.305/4.788 & --- \\
UniAD~\cite{ref-uniad}\textsuperscript{\dag} & single, zero-shot & 202 kf & 1.906/3.428 & 3.580/6.735 & 5.246/10.141 & --- \\
\end{longtable}
\endgroup

On the same 202-keyframe population, the fine-tuned checkpoint
outperformed the zero-shot baseline at every horizon. UniAD and AutoE2E
both adopt a BEV-centric representation and planning-oriented feature
sharing, but UniAD unifies perception, tracking, mapping, motion
forecasting, occupancy, and planning through task interaction, whereas
AutoE2E narrows its scope to temporal camera BEV, explicit navigation
context, auxiliary route reconstruction, and direct ego planning.
Table~\ref{tab-crossmodel} now reports UniAD on the same 202-keyframe
population as AutoE2E, but only zero-shot, with no KITScenes-specific
retraining, against UniAD's native nuScenes training; the roughly
five-fold gap in 3\,s FDE this produces against AutoE2E's fine-tuned
checkpoint is read as a domain-shift effect, not a capability claim
against UniAD's architecture, since no comparison under the same
zero-shot/fine-tuned distinction applied to AutoE2E has been run on
UniAD's own nuScenes benchmark. UniAD was evaluated on more than one
KITScenes population -- including the full 11,035-sample dense
population, documented alongside the 202-keyframe run in
\texttt{Model/accuracy\_benchmark/uniad/} -- but only the 202-keyframe
numbers are presented here, consistent with this comparison's scope.

The evidence in this paper does not support a causal claim that the HD
map and the route improved planning accuracy: Val and Test are
different scenes, Test by construction has no map and no route, this
report contains no camera-only baseline on the same scenes as Val, and
per-scene paired errors have not been computed. What can be said is
behavioral: route-corridor compliance (the fraction of steps a 2.0\,m
footprint remains inside the 3.5\,m route corridor) is 0.40--0.42 and
success 0.06--0.08 under this strict definition, and a route
reconstruction IoU of 0.999 verifies that the route information
survives the gate, without being evidence the policy drove along the
route.

\subsection{Qualitative Examples}\label{sec:qualitative}

\begin{figure}[h]
\centering
\includegraphics[width=0.8\linewidth]{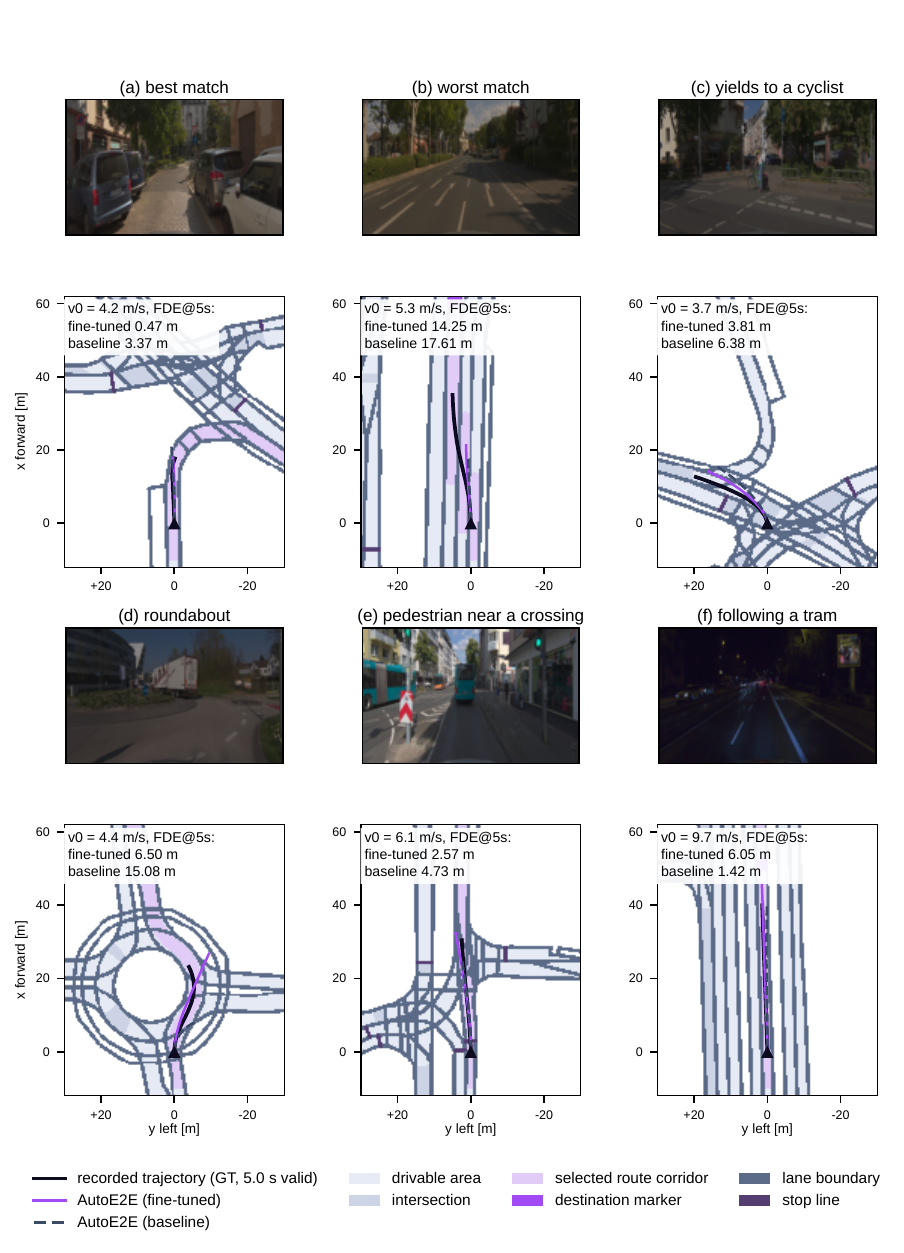}
\caption{Qualitative examples from KITScenes. The top row shows the
model-input front tile of each sample (center crop); the bottom row
shows the logged trajectory (obsidian, 5\,s) and the replayed
trajectories of the fine-tuned and baseline checkpoints on the sample's
own map and route raster. The 5\,s FDE is annotated in each
panel.}\label{fig-qual}
\end{figure}

Fig.~\ref{fig-qual} shows six KITScenes examples replayed with the
unicycle model over each sample's own map and route raster, pulled live
from the DataModelConsole dashboard (Sec.~\ref{sec:dmc}): the closest
and the furthest match to the recorded path, yielding to a cyclist, a
roundabout, a pedestrian near a crossing, and following a tram --
chosen for behavioral variety over a repeated empty-road scene, since a
straight, empty road is not informative about planning behavior. Unlike
the aggregate Val/Test tables above, each panel shows the actual road
and route context (drivable area, lane boundaries, the selected route
corridor) alongside the camera view, so a reader can see what the model
saw. In the roundabout example, the fine-tuned checkpoint's error is
6.50\,m against the baseline's 15.08\,m; in the worst-case example, both
checkpoints overshoot the logged trajectory because neither predicts
the driver's deceleration. These are single-sample observations, not
statistical conclusions.

\begin{figure}[h]
\centering
\includegraphics[width=0.8\linewidth]{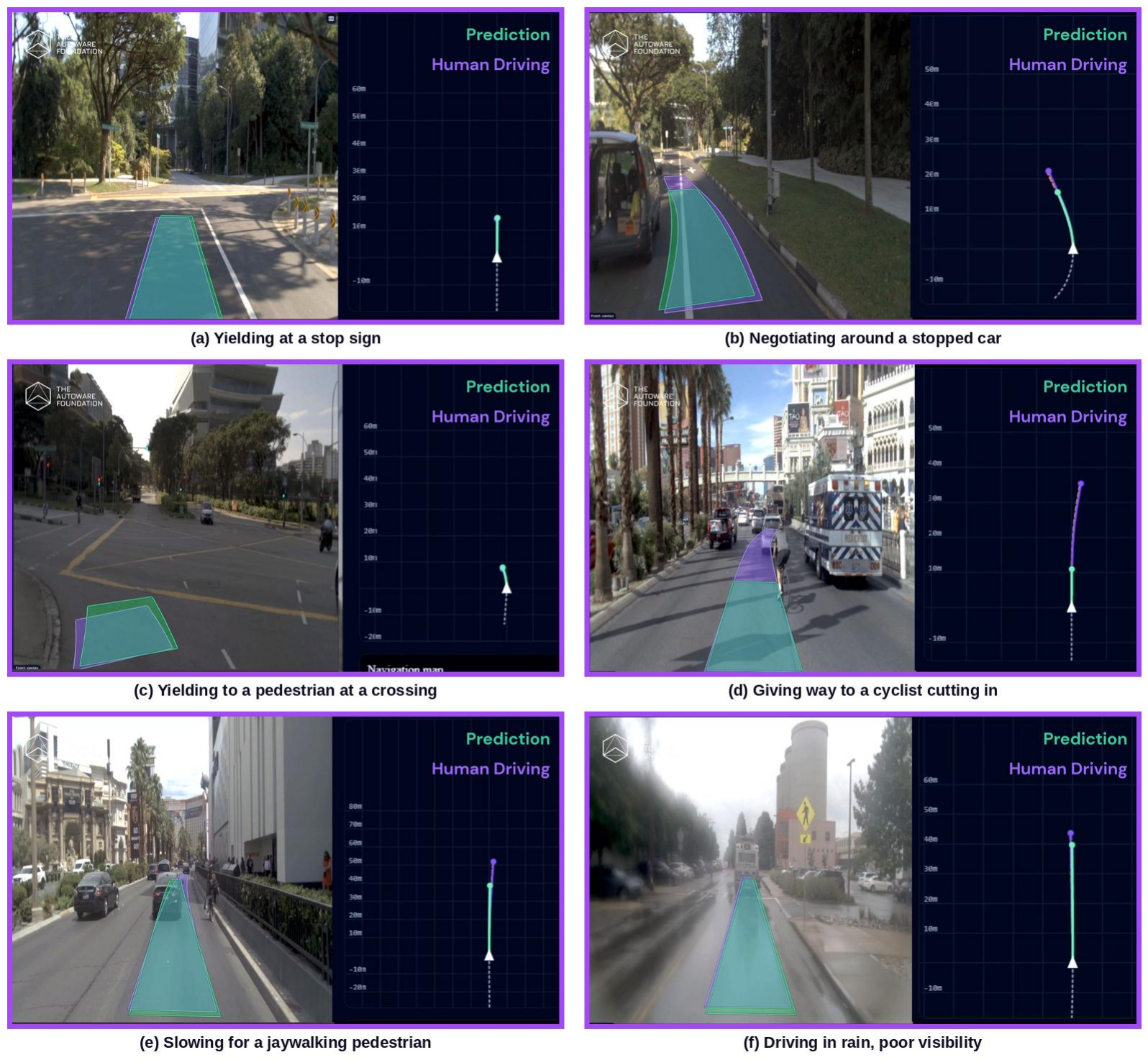}
\caption{Real-world qualitative behavior captures from the
DataModelConsole dashboard (official Autoware Foundation Robotaxi
Working Group material). Teal is AutoE2E's predicted trajectory, violet
the recorded human driving, overlaid on the front camera view and a
bird's-eye-view panel.}\label{fig-console}
\end{figure}

Beyond these benchmark-tied KITScenes examples, Fig.~\ref{fig-console}
shows six real-world behavior captures recorded with the
DataModelConsole dashboard (Sec.~\ref{sec:dmc}), spanning driving
locations outside the KITScenes cities of
Sec.~\ref{sec:datasets} -- evidence that the behaviors below are not an
artifact of one city's road layout. Each panel overlays AutoE2E's
predicted trajectory (teal) on the camera view and a bird's-eye-view
panel against the recorded human driving (violet): slowing for and
avoiding a jaywalking pedestrian, giving way to a cyclist cutting in,
negotiating around a stopped car, yielding to a pedestrian at a
crossing, yielding at a stop sign, and driving in rain and poor
visibility. These are curated, single-scene captures selected by the
working group to illustrate specific behaviors, not a random or
representative sample, and are reported as qualitative illustration
alongside, not evidence for, the quantitative claims in
Sec.~\ref{sec:results-val}--\ref{sec:cross-model}.

\subsection{Inference Cost and Real-Vehicle Evidence}\label{sec:realvehicle}

Across a sweep of backbone and fusion configurations, inference reaches
up to \textbf{76\,FPS} with a SwinV2-Tiny backbone and feature-concat
fusion (RTX\,5080, batch 1) -- a lighter-weight variant benchmarked for
throughput, not the ResNet-50 configuration evaluated in
Sec.~\ref{sec:benchmarks}. Both configurations have been benchmarked
across consumer and workstation GPUs (RTX\,5000 Quadro, A40,
RTX\,4070/4080/5090); a controlled latency measurement of the exact
evaluated ResNet-50 configuration on this same GPU sweep is not
available in the evidence used for this paper. The only concrete
latency figure for that configuration comes from TU Munich's EDGAR
platform (Sec.~\ref{sec:tum}): 101\,ms on a single A6000 GPU, for a
checkpoint TUM retrained on its own EDGAR data rather than the
KITScenes-trained checkpoints benchmarked above -- reported there as a
preliminary, single-platform perspective, not a closed validation.

\section{Limitations}\label{sec:limitations}

This paper documents an early snapshot of an actively developed
reactive band, released at this stage so the working group's evaluation
protocol and baseline evidence are open to scrutiny and reuse rather
than held back until every gap below is closed; Sec.~\ref{sec:future}
describes the concrete improvements already planned. AutoE2E is
currently an open-loop model: it is not designed as a closed-loop
policy, has not been evaluated under closed-loop safety or intervention
metrics, and future versions are intended to be trained and evaluated
on real vehicles directly rather than assumed safe to deploy on the
basis of the open-loop numbers in this paper.

The evaluation in this paper is open-loop only and includes no
closed-loop safety or intervention metrics. No controlled map/route
experiment on identical scenes was conducted. The external replay is
limited to 5\,s, so 22\% of the 6.4\,s output is not evaluated. Val and
Test are different populations, and the difference between them carries
no causal meaning. The camera encoder is frozen, and the degree to
which a representation trained on nuScenes fits the six-camera
configuration and 0.6\,m grid of KITScenes has not been separately
evaluated. The full UniAD task stack is not included. All checkpoint
quality gates are recorded as not passed. The training data are nuPlan
and KITScenes from a small number of German and multi-city scenes, so
distribution shifts in geography, season, and map may exist. Because
the route is a post-hoc sequence of already-driven lanes, robustness to
route errors is not measured. Under the KITScenes benchmark protocol,
only 4.0\,s of the 6.4\,s ego-motion history is real data, leaving a
distribution gap with respect to training.

Deployment-aligned evaluation should retain the navigation inputs the
policy requires, camera-only evaluation should be positioned as a
complementary robustness track rather than the primary benchmark, and
the real-vehicle and cross-model results of this paper should be read
as preliminary evidence pending the full same-scene ablation, not as
settled claims.

\section{Future Work and Outlook}\label{sec:future}

AutoE2E is the reactive band of a three-band architecture planned for
the Autoware Foundation's open-source autonomous-driving stack: a
reactive band (this paper), a reasoning band, and a world-action band,
decoupled rather than trained as one monolithic network. This paper is
explicitly scoped to the frozen-camera-backbone version of AutoE2E
described in Sec.~\ref{sec:architecture}; unfreezing the camera BEV
pathway for joint fine-tuning is left to future work and is not
evaluated here. Concretely, the working group's roadmap includes:

\begin{itemize}[leftmargin=1.4em, itemsep=2pt]
  \item \textbf{Unfreezing the camera BEV backbone} for joint fine-tuning
  with the navigation and control pathways, expected to require its own
  evaluation protocol rather than a direct comparison with the
  frozen-backbone results reported here.
  \item \textbf{Replacing the HD-map dependency with a navigation map.}
  The current deployment contract requires an HD map; a navigation- or
  SD-map-conditioned variant would relax this requirement toward the
  maps actually available at scale.
  \item \textbf{Closed-loop evaluation and training} in a photorealistic
  world model, rather than the open-loop imitation setting evaluated
  throughout this paper. A first closed-loop integration with NVIDIA's
  AlpaSim platform has been run: the baseline checkpoint completed five
  public NuRec scenes without collision, but left the road in four of
  five within 2.5--4.1\,s of takeover. A replay ablation traced this to
  the model's own yaw-rate history feeding back closed-loop (a
  ``copycat'' failure also seen in other imitation-learned policies, not
  a configuration bug) -- zeroing that history at replay keeps the
  logged plan within about 2$^\circ$ of straight even where the live
  rollout had already drifted 5--10$^\circ$ off-lane. The next steps are
  a closed-loop run with that history zeroed at inference, and training
  with less dependence on it (dropout, noise, or a shorter history).
  \item \textbf{Activating the reasoning and world-action bands},
  Fig.~\ref{fig-threeband}'s JEPA-style future-feature prediction
  included, so that the reactive policy evaluated here can be coupled
  with a higher-level reasoning band and a learned world model rather
  than being extended into a single larger network.
   \item \textbf{Real-world vehicle deployment}, extending the
  preliminary results on TUM's EDGAR platform (Sec.~\ref{sec:tum}) to
  closed-loop driving on KIT's test vehicle \emph{Joy} in German
  inner-city traffic (Sec.~\ref{sec:kit}), toward a road-ready rather
  than demo-grade system.
\end{itemize}

\section{Collaborations}\label{sec:collaboration}

\begin{partnerbox}{Technical University of Munich (TUM) -- EDGAR Research Platform}
\label{sec:tum}
\noindent
\begin{minipage}[t]{0.68\linewidth}
A preliminary real-vehicle result, coordinated with the working group
with TUM's EDGAR team, is reported here as a
perspective on where this work goes next -- not a closed validation.
The full experimental methodology, including whether every compared
model received the same adaptation to EDGAR data, has not yet been
independently confirmed with TUM, and the result is single-platform;
read it accordingly. Concretely:
\end{minipage}%
\hfill
\begin{minipage}[t]{0.28\linewidth}
\vspace{0pt}
\includegraphics[width=0.8\linewidth]{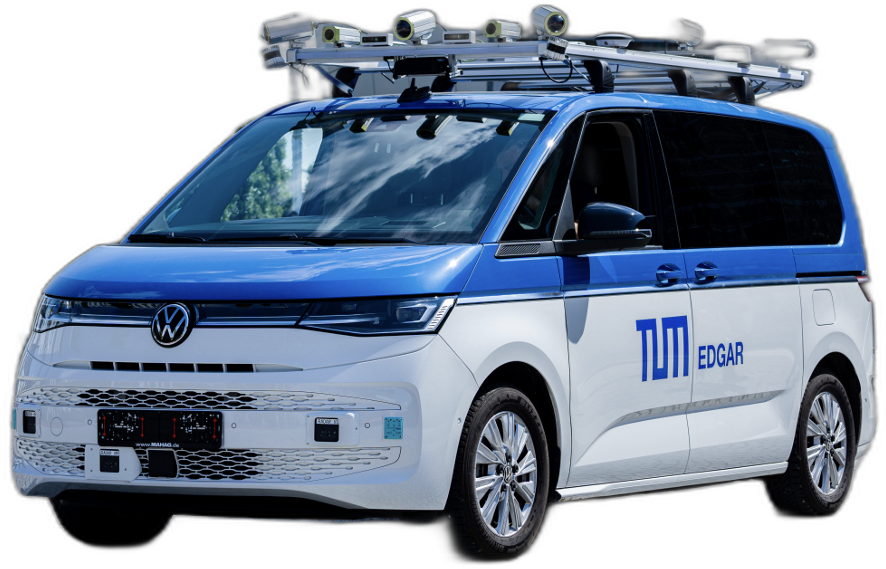}
\end{minipage}
\begin{itemize}[leftmargin=1.4em, itemsep=2pt]
  \item \textbf{Real-vehicle retraining:} TUM trained AutoE2E from
  scratch on 134 shards ($\sim$27{,}000 samples) of its own EDGAR
  vehicle data, reaching 0.430\,m ADE and 1.224\,m FDE at 3\,s -- against
  a constant-velocity reference of 0.955\,m / 2.605\,m on the same data
  -- within a few hours of training by the working group's account.
  \item \textbf{Efficiency evidence:} inference on a single A6000 GPU is
  101\,ms at 0.5\,GiB memory. Compared zero-shot on the same EDGAR data,
  Alpamayo\,1.5 reaches $\sim$0.99\,m (segment 1) and $\sim$0.64\,m
  (segment 2) ADE at 3\,s, and SimForge-D1.5 reaches $\sim$1.00\,m and
  $\sim$0.76\,m across the same two segments, both at roughly 800\,ms
  inference and 14--21\,GiB memory -- the first evidence that AutoE2E's
  parameter budget and architecture translate into a real latency and
  memory advantage on vehicle hardware, not only on the open-loop
  benchmarks of Sec.~\ref{sec:benchmarks}.
\end{itemize}
\end{partnerbox}

\begin{partnerbox}{Karlsruhe Institute of Technology (KIT) -- Joining the Autoware Foundation}
\label{sec:kit}
\noindent
\begin{minipage}[t]{0.68\linewidth}
KIT, one of Germany's leading technical universities, has joined the
Autoware Foundation as an Academic Member, and its CITI Drive Team
at the Institute of Measurement and Control Systems (MRT) will
collaborate with Autoware to bring full-stack end-to-end driving onto
European roads. KIT supports Autoware's Robotaxi Working Group in its
efforts towards E2E driving:
\end{minipage}%
\hfill
\begin{minipage}[t]{0.28\linewidth}
\vspace{0pt}
\includegraphics[width=\linewidth]{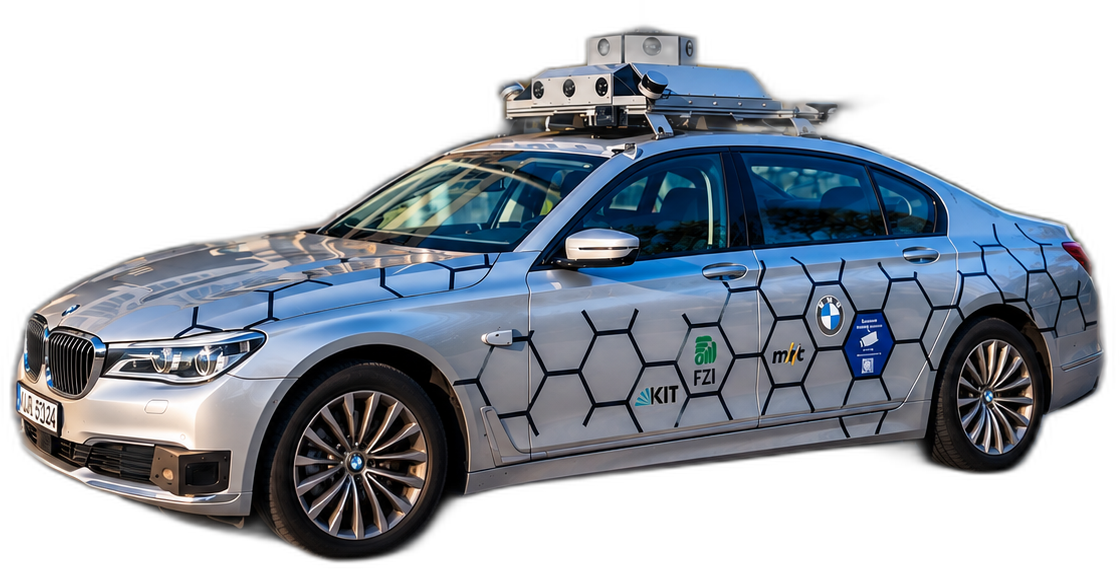}
\end{minipage}
\begin{itemize}[leftmargin=1.4em, itemsep=2pt]
  \item \textbf{Independent benchmarking:} KIT is independently
  benchmarking Autoware's open-loop trained model release on
  KITScenes-Multimodal -- the urban autonomous driving dataset combining
  high-fidelity sensor data with complete Lanelet2 HD maps.
  \item \textbf{Training and real-world deployment:} KIT will
  collaborate with Autoware to further train and deploy the E2E model on
  its test vehicle \emph{Joy} and to demonstrate autonomous driving in
  Germany.
\end{itemize}
\textbf{Autoware E2E on European roads.} \emph{Joy} will serve as the
main target platform for Autoware E2E in Europe. Having recorded
KITScenes in Karlsruhe, Frankfurt, and Sindelfingen, \emph{Joy} takes
end-to-end driving onto challenging German inner-city roads -- from
dataset to closed-loop road deployment with minimal domain gap.
\end{partnerbox}

Beyond this paper, the working group's own publication plan continues
with a camera-focused follow-up, a tools-and-evaluation paper (in
discussion with KIT, which could also give the working group visibility
on KIT's own benchmark page), and a fused-model, closed-loop paper once
the roadmap items above land.

\section{Contributors and Acknowledgments}\label{sec:contributors}

Contributors are listed in three tiers, following the working group's
own contact records: Project Leads, Core Contributors (split between
architecture and benchmarking, people may appear in both), and Other
Contributions from the Karlsruhe Institute of Technology and TU Munich.

\subsection{Project Leads}
Muhammad Zain Khawaja: President of the Autoware Foundation, Robotaxi
Working Group direction, architecture design, benchmark design and
review. Zaynab El Mawas: Co-chair Robotaxi Working Group, research
lead, paper preparation and working-group coordination (University of
Lille, France).

\subsection{Core Contributors}
\begin{itemize}
    \item \textbf{Architecture:} Ryota Yamada (Amazon Web Services, Japan) --
architecture, training, evaluation, and code and repository
maintenance; Gabriela Cordova (Universidad Carlos III de Madrid, Spain); Arseni Lysak
(Warsaw University of Technology, Poland); Intisar Chowdhury (SENSYN
ROBOTICS, Inc., Japan); Bharat Jayaprakash (University of Minnesota,
USA); Apostolos Georgiadis (nfiniity GmbH, Germany); Shri Harsha Adapala
(University at Buffalo, USA); Jatin Gautam (Independent researcher);
Christoph Bank (TU Munich, Germany); and Alejandra Solarte Uscátegui
(Tecnocad, Italy).
\item \textbf{Benchmarking:} Ryota Yamada -- AutoE2E/nuPlan on KITScenes; Shri
Harsha Adapala -- SimForge; Atanasko Boris Mitrev (EXO Service
Solutions, North Macedonia) -- Qwen-Drive-1.0, Alpamayo\,1.5; Hassan Ali
Ansari (TRATON Group R\&D, Sweden) -- Drive-JEPA; Nishad Sahu (Carnegie
Mellon University, USA) -- UniAD; Arseni Lysak -- METEOR.
\end{itemize}

\subsection{Other Contributions}
\begin{itemize}
    \item \textbf{TU Munich (TUM):} Christian Oefinger, coordinating the EDGAR team's real-vehicle work. (Sec.~\ref{sec:tum}).
    \item \textbf{Karlsruhe Institute of Technology (KIT-MRT):} its
KITScenes team (Richard Schwarzkopf, Fabian Immel) for the
dataset~\cite{ref-kitscenes} underlying this paper's evaluation, and
its CITI Drive team (Piotr Spieker, Richard Schwarzkopf) for the
upcoming real-road deployment on \emph{Joy} (Sec.~\ref{sec:kit}).
\end{itemize}

\subsection*{Acknowledgments}
We thank the contributors and the community of the Autoware Foundation
Robotaxi Working Group; Ryohsuke Mitsudome, Simon Thompson, Bin Wang,
and Shunya Seiya of TierIV for their support and collaboration.

\bibliographystyle{unsrt}

\end{document}